%% file: main.tex
\documentclass{article}

\usepackage{microtype}
\usepackage{graphicx}
\usepackage{subfigure}
\usepackage{booktabs} % for professional tables

\usepackage{hyperref}

\usepackage[accepted]{icml2025}
\usepackage{booktabs} % For professional quality tables
\usepackage{tabularx} % To create tables with columns that automatically wrap text
\usepackage{amsmath}
\usepackage{amssymb}
\usepackage{mathtools}
\usepackage{amsthm}
\usepackage{float}
\usepackage{subcaption}
\usepackage{enumitem}
\usepackage{placeins}
\usepackage{booktabs}   % 用于创建更美观的横向表格线
\usepackage{ragged2e}
\usepackage{multirow}   % 用于创建跨越多行的单元格
\usepackage{colortbl}   % 用于给单元格或行添加背景色
\usepackage{graphicx}   % 如果您想旋转文字（这里没用，但备用）
\usepackage{subcaption}
\usepackage[table]{xcolor} % 提供更灵活的颜色定义，colortbl 依赖于此
\usepackage[most]{tcolorbox}
\usepackage{pifont}

\usepackage[capitalize,noabbrev]{cleveref}

\theoremstyle{plain}

\theoremstyle{definition}

\theoremstyle{remark}

\usepackage[textsize=tiny]{todonotes}
\usepackage{xltabular}
\usepackage{booktabs}
\newcommand{\MYMETHOD}{AlphaPROBE}

\icmltitlerunning{\MYMETHOD: \textbf{Alpha} Mining via \textbf{P}rincipled \textbf{R}etrieval and \textbf{O}n-graph \textbf{B}iased \textbf{E}volution}

\begin{document}

\twocolumn[
\icmltitle{\MYMETHOD: \textbf{Alpha} Mining via \textbf{P}rincipled \textbf{R}etrieval and \textbf{O}n-graph \textbf{B}iased \textbf{E}volution}



\icmlsetsymbol{leader}{\ensuremath{\dagger}}
\icmlsetsymbol{corr}{\ding{41}}

\begin{icmlauthorlist}
\icmlauthor{Taian Guo}{pkucs,comp}
\icmlauthor{Haiyang Shen}{pku,leader}
\icmlauthor{Junyu Luo}{pkucs}
\icmlauthor{Binqi Chen}{pkucs,comp}
\icmlauthor{Hongjun Ding}{sch}

\icmlauthor{Jinsheng Huang}{pkucs,comp}
\icmlauthor{Luchen Liu}{comp}
\icmlauthor{Yun Ma}{pku,corr}
\icmlauthor{Ming Zhang}{pkucs,corr}
\end{icmlauthorlist}

\icmlaffiliation{pkucs}{National Key Laboratory for Multimedia Information Processing, School of Computer Science, PKU-Anker LLM Lab, Peking University}
\icmlaffiliation{pku}{Institute for Artificial Intelligence, Peking University}
\icmlaffiliation{comp}{Zhengren Quant, Beijing, China}
\icmlaffiliation{sch}{Baruch College, City University of New York}

\icmlcorrespondingauthor{Yun Ma}{mayun@pku.edu.cn}
\icmlcorrespondingauthor{Ming Zhang}{mzhang\_cs@pku.edu.cn}

\icmlkeywords{Alpha Mining, Large Language Models, Principled Retrieval}

\vskip 0.3in
]



\printAffiliationsAndNotice{\textsuperscript{\ensuremath{\dagger}}Project Leader. \textsuperscript{\ding{41}}Corresponding. Contact: Taian Guo \textless{}taianguo@stu.pku.edu.cn\textgreater{}, Haiyang Shen \textless{}hyshen@stu.pku.edu.cn\textgreater{}.}

\definecolor{highlightcolor}{gray}{0.9}

\begin{abstract}

Extracting signals through alpha factor mining is a fundamental challenge in quantitative finance. Existing automated methods primarily follow two paradigms: Decoupled Factor Generation, which treats factor discovery as isolated events, and Iterative Factor Evolution, which focuses on local parent-child refinements. However, both paradigms lack a global structural view, often treating factor pools as unstructured collections or fragmented chains, which leads to redundant search and limited diversity. To address these limitations, we introduce \MYMETHOD~(\textbf{Alpha} Mining via \textbf{P}rincipled \textbf{R}etrieval and \textbf{O}n-graph \textbf{B}iased \textbf{E}volution), a framework that reframes alpha mining as the strategic navigation of a Directed Acyclic Graph (DAG). By modeling factors as nodes and evolutionary links as edges, \MYMETHOD~treats the factor pool as a dynamic, interconnected ecosystem. The framework consists of two core components: a Bayesian Factor Retriever that identifies high-potential seeds by balancing exploitation and exploration through a posterior probability model, and a DAG-aware Factor Generator that leverages the full ancestral trace of factors to produce context-aware, non-redundant optimizations. Extensive experiments on three major Chinese stock market datasets against 8 competitive baselines demonstrate that \MYMETHOD~significantly gains enhanced performance in predictive accuracy, return stability and training efficiency. Our results confirm that leveraging global evolutionary topology is essential for efficient and robust automated alpha discovery. We have open-sourced our implementation at ~\url{https://github.com/gta0804/AlphaPROBE}.

\end{abstract}

% \vspace{-1em}

\input{sections/Introduction}
\input{sections/RelatedWork}
\input{sections/Methodology}
\input{sections/Evaluation}

\input{sections/Conclusion}


\bibliography{main}
\bibliographystyle{icml2025}

%%%%%%%%%%%%%%%%%%%%%%%%%%%%%%%%%%%%%%%%%%%%%%%%%%%%%%%%%%%%%%%%%%%%%%%%%%%%%%%
%%%%%%%%%%%%%%%%%%%%%%%%%%%%%%%%%%%%%%%%%%%%%%%%%%%%%%%%%%%%%%%%%%%%%%%%%%%%%%%
% APPENDIX
%%%%%%%%%%%%%%%%%%%%%%%%%%%%%%%%%%%%%%%%%%%%%%%%%%%%%%%%%%%%%%%%%%%%%%%%%%%%%%%
%%%%%%%%%%%%%%%%%%%%%%%%%%%%%%%%%%%%%%%%%%%%%%%%%%%%%%%%%%%%%%%%%%%%%%%%%%%%%%%
\newpage
\appendix
\input{sections/Appendix}

\end{document}

%% file: sections/Introduction.tex
\section{Introduction}

Extracting predictive signals from noisy and high-dimensional market data is a fundamental challenge in quantitative finance~\cite{cuthbertson2005quantitative,lee2010handbook,wilmott2013paul,rundo2019machine,sun2023reinforcement}. The primary approach to this problem is \textit{alpha factor mining}. This process involves discovering mathematical expressions, known as alpha factors, that transform raw market data into predictors of future asset returns. Rather than searching for a single perfect predictor, the objective is to build a diverse portfolio of factors that provide collective and robust predictive power. To achieve this, several automated discovery methods have recently emerged.

These automated methods generally follow two paradigms. The first is \textit{Decoupled Factor Generation (DFG)}, where models generate factors independently based on a shared training distribution~\cite{yu2023alphagen, zhao2025quantfactor, alphaqcm, chen2025alphasage, shi2025alphaforge, trajectorylevel}. In this paradigm, the relationship between factors remains implicit and weak because the model treats each generation attempt as an isolated event. The second paradigm is \textit{Iterative Factor Evolution (IFE)}, which focuses on refining existing factors into improved descendants~\cite{alphagpt, li2024FAMA, luo2025efs, tang2025alphaagent, shi2025navigating, li2025rdagentquant, cao2025chainoflapha}. While IFE considers the immediate parent of a factor, it typically optimizes local chains and overlooks the global relationship between all factors in the pool.

The lack of a global structural view represents a critical limitation shared by both paradigms. DFG methods treat the growing factor library as an unstructured collection and ignore the connections between similar expressions. Meanwhile, IFE methods view evolution through a narrow lens and focus on single parent-child pairs instead of the broader evolutionary web. Consequently, existing methods prioritize the quality of individual factors but overlook the strategic information embedded in their topological relationships. We contend that modeling these relationships is a core requirement for an efficient discovery process. However, organizing factors into a structure is only the beginning. The primary challenge is how to navigate this complex topology to guide future discovery.

To address these challenges, we introduce \MYMETHOD. This framework reframes alpha mining as the strategic navigation of a Directed Acyclic Graph (DAG). Our design is motivated by the need to treat the factor pool as a dynamic and interconnected ecosystem rather than a static list. By representing factors as nodes and their evolutionary links as edges, we can apply structural reasoning to the discovery process. We implement this vision through a closed-loop system comprising two core components: the \textit{Bayesian Factor Retriever} and the \textit{DAG-aware Factor Generator}. These components work together to ensure the search for new factors is efficient and structurally informed.

The Bayesian Factor Retriever is designed to solve the exploration-exploitation trade-off during parent selection. It evaluates each factor in the DAG based on its potential using a posterior probability. The prior term in this model accounts for exploitation by rewarding high-performing factors. It also penalizes factors that are over-optimized or already frequently used. Simultaneously, the likelihood term drives exploration by assessing how much novel information a potential descendant might contribute to the existing pool. By analyzing the position of a factor in the global topology, the Retriever identifies the most promising seeds for the next generation of alpha factors.

Once the best parent factors are selected, the DAG-aware Factor Generator performs targeted optimizations to create new candidates. Unlike standard generators that only look at a single parent expression, our LLM-based component leverages the entire generation trace. This trace represents the full ancestral path from the root node to the selected parent. This historical context allows the LLM to understand which modification strategies have already been tried and which remains unexplored. This approach prevents redundant mutations and encourages the generation of more diverse factors. These new factors are then integrated back into the DAG, which completes the evolutionary cycle.

We conduct extensive experiments to validate \MYMETHOD~on three major Chinese stock market datasets, including CSI 300, 500, and 1000. \MYMETHOD~is compared against 8 competitive baselines from three distinct categories. These include expert-designed pools, reinforcement learning methods, and state-of-the-art LLM-based agents. The results demonstrate that \MYMETHOD~outperforms existing methods in predictive accuracy and return stability. Furthermore, we provide detailed ablation studies to verify our Bayesian retrieval mechanism. We also conduct parameter sensitivity analyses to confirm the effectiveness of DAG-based generation and a detailed case study to inspect the evolutionary path of generated factors. These experiments prove that leveraging evolutionary structure is superior for automated alpha discovery.

Our main contributions are summarized as follows:

\begin{itemize}[left=0.3cm,partopsep=0pt,topsep=-2pt,itemsep=2pt]
    \item We introduce \MYMETHOD, a framework that models alpha factor evolution as a DAG. Our goal is to move beyond unstructured factor collections by capturing global relationships. This topological view enables more strategic navigation of the massive factor search space.

    \item We develop a closed-loop discovery system using a Bayesian Factor Retriever and a DAG-aware Factor Generator. The retriever identifies high-potential seeds by balancing quality and diversity through a posterior model. To avoid redundant mutations, the generator utilizes the full ancestral lineage to produce context-aware and novel factor optimizations.

    \item We validate \MYMETHOD~on three major stock datasets against 8 SOTA baselines. The experiments demonstrate that \MYMETHOD~significantly improves predictive performance, return stability and training efficiency. Our analysis also confirms that this approach creates a more robust and hierarchically organized factor library.
\end{itemize}

%% file: sections/RelatedWork.tex
\section{Background}

\begin{figure}[htbp!]
    \centering
    \includegraphics[width=0.7\linewidth]{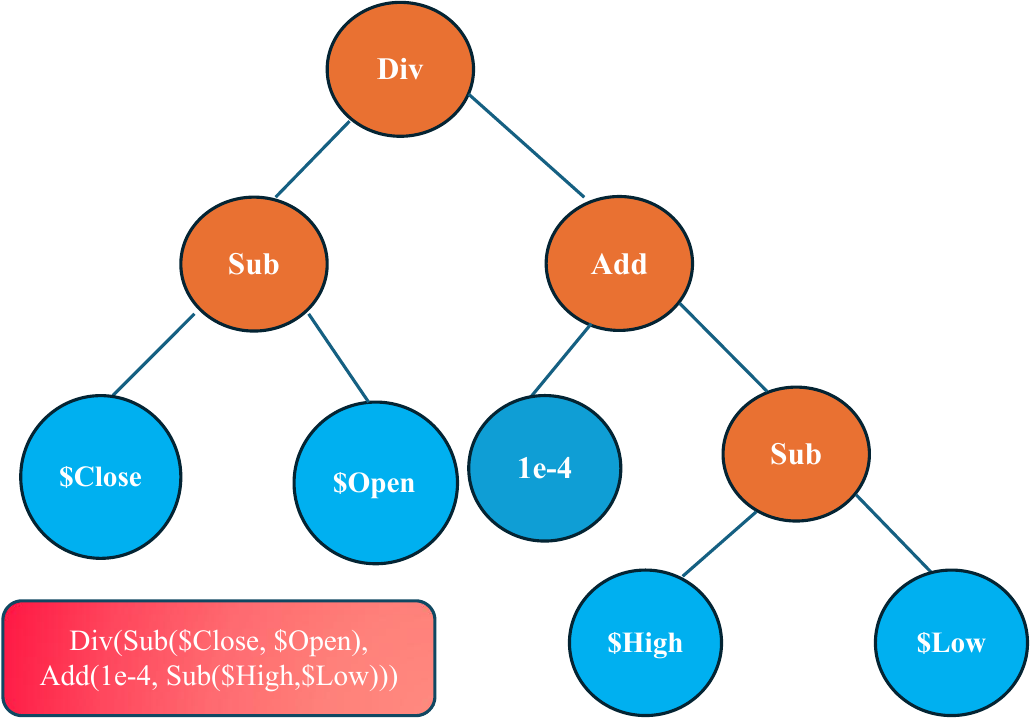}
    \caption{An example of alpha factor measuring the normalized daily price change on the Abstract Syntax Tree(AST) and expression view. All available features and operators are in the Appendix ~\ref{appendix::detail:featop}.}
    \label{fig:factorExample}.
\end{figure}

In trading, a factor is a predictive signal derived from various data sources. The input data for an asset at time $t$ typically includes price-volume features like open ($O_t$), high ($H_t$), low ($L_t$), close ($C_t$), and volume ($V_t$). Beyond these, modern systems also incorporate fundamental data, such as earnings and cash flow from financial reports, as well as alternative data like news sentiment. An alpha factor is a mathematical expression $f$ that transforms these data streams into a single score. As shown in Figure~\ref{fig:factorExample}, a factor is usually represented as an expression tree. In this tree, leaf nodes represent input features, while internal nodes represent mathematical operators like \texttt{Add}, \texttt{Sub}, or \texttt{TsMean}.

The core objective of alpha mining is to discover an optimal set of factors $\mathcal{A} = \{f_1, f_2, \dots, f_N\}$. These factors are selected from a massive search space $\mathcal{F}$, which contains all possible valid mathematical combinations. We formalize this task as a maximization problem of a portfolio-level utility function $\mathcal{J}(\mathcal{A})$:

\begin{equation}
\max_{\mathcal{A} \subset \mathcal{F}} \mathcal{J}(\mathcal{A})
\end{equation}

The function $\mathcal{J}$ measures the overall quality of the factor pool. In practice, $\mathcal{J}$ represents specific financial goals. For example, it can be the Sharpe Ratio of a portfolio built from these factors, which measures the return per unit of risk. Alternatively, it can be a combination of the Information Coefficient (IC) for predictive accuracy and a penalty term for inter-factor correlation to ensure diversity. By maximizing $\mathcal{J}$, we ensure that the discovered factors are powerful and complementary. This formalized goal has led to the development of various automated discovery methods, which we categorize based on their generation logic.

\section{Related Work}

\subsection{Decoupled Factor Generation (DFG)}

The first paradigm, Decoupled Factor Generation (DFG), focuses on creating factors as independent individual units. These methods typically use Reinforcement Learning~\cite{ppo, yu2023alphagen, alphaqcm, jiang2025alpha}, Graph Neural Networks~\cite{alphagat}, or Generative Models~\cite{shi2025alphaforge, gan} to sample factors from a vast search space. Recent works have also explored enhancing sample diversity via Generative Flow Networks~\cite{chen2025alphasage} or multi-distribution learning~\cite{zhao2025quantfactor, trajectorylevel}. In this paradigm, models are trained on shared data, allowing for implicit knowledge sharing. However, the generation process treats each attempt as an isolated event. Because the model ignores the specific state of the current library, the relationships between factors remain implicit and weak.

While DFG methods excel at broad exploration, their unstructured approach limits the ability to build upon previous discoveries systematically. By treating the factor pool as a simple collection rather than a connected system, these methods ignore the evolutionary logic that leads to success. This gap suggests a need for methods that can refine existing knowledge through iterative improvements. This realization leads us to the second major paradigm in the field.

\subsection{Iterative Factor Evolution (IFE)}

The second paradigm, Iterative Factor Evolution (IFE), treats factor discovery as a process of continuous refinement. These methods start with a ``parent'' factor and apply modifications to create improved ``children''. Early methods in this category used Genetic Programming to evolve expression trees through crossover and mutation~\cite{zhang2020autoalpha, chen2021gp}. Recently, Large Language Models (LLMs) have been used to provide intelligent feedback for these refinements~\cite{alphagpt, li2024FAMA, luo2025efs}. Modern agents now use backtesting results to prompt LLMs for targeted formula changes~\cite{tang2025alphaagent, cao2025chainoflapha, li2025rdagentquant}. Some frameworks even use multi-agent systems~\cite{zhi2025automate, liu2025cognitive} or Monte Carlo Tree Search~\cite{browne2012survey, shi2025navigating} to guide local search.

The primary limitation of current IFE methods is their lack of a global view. By focusing only on single-parent-child pairs or linear chains, these methods overlook the broader evolutionary web of the entire factor pool. They lack a principled mechanism to choose the most promising seed from a large library of candidates. Consequently, the discovery process often becomes repetitive or trapped in local optima. To solve these problems, \MYMETHOD~organizes the factor pool into a DAG to enable global and strategic navigation.

%% file: sections/Methodology.tex
\section{Method}

We reframe the alpha factor mining problem as a strategic navigation and creation process on a DAG $\mathcal{G} = (\mathcal{F}, \mathcal{E})$, where $\mathcal{F}$ is the set of all discovered factors (nodes) and $\mathcal{E}$ is the set of directed edges representing their lineage. An edge $(F_p, F_c) \in \mathcal{E}$ indicates that factor $F_c$ was derived from parent $F_p$. Within this framework, the overarching optimization problem bifurcates into two distinct, interconnected challenges:

\begin{itemize}[left=0.3cm,partopsep=0pt,topsep=-2pt,itemsep=-1pt]
    \item \textbf{Strategic Retrieval}: The first challenge is to identify which existing factor in the graph is the most promising candidate for evolution. This is a search problem on the DAG to find an optimal parent factor $F_p^*$ that maximizes the expected quality of its yet-to-be-created descendant, $F_{new}$. Formally, we seek to solve:
    \begin{equation}
        F_p^* = \arg\max_{F_p \in \mathcal{F}} \mathbb{E}[ \text{Qual}(F_{new}) | \text{parent}(F_{new})=F_p, \mathcal{G} ]
        \vspace{-0.5em}
    \end{equation}
    where $\text{Qual}(\cdot)$ is a measure of factor quality.
    \item \textbf{Targeted Generation}: The second challenge is to generate novel, high-quality offspring from the selected parent. This is a generation problem that leverages the parent's contextual information within the graph—specifically, its ancestral path—to guide the creation of new $\{F_c\}$. This process can be defined as:
    \begin{equation}
        \{F_{c,1}, \dots, F_{c,k}\} = G(F_p^*, T(F_p^*))
    \end{equation}
    where $\{F_{c,1}, \dots, F_{c,k}\}$ is a newly generated child factor, $T(F_p^*)$ is the generation trace (the path from a root to $F_p^*$ in $\mathcal{G}$), and $G$ is the generation function.
\end{itemize}

\begin{figure*}[t]
  \centering
  \includegraphics[width=0.80\linewidth]{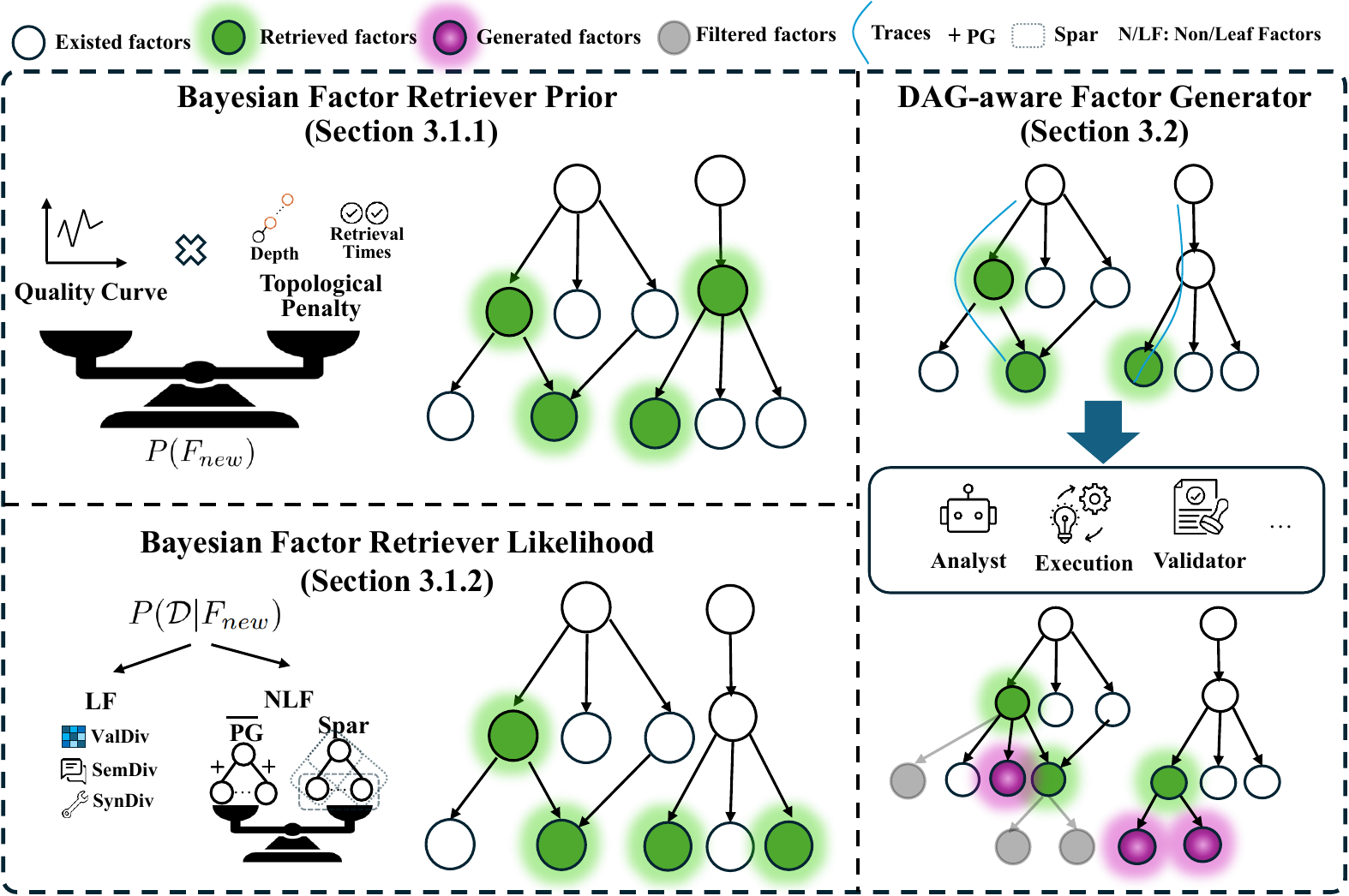}
  \caption{The overall illustration of \MYMETHOD. ~\MYMETHOD~ is a closed-loop framework consisting of a Bayesian Factor Retriever and a DAG-aware Factor Generator. The Bayesian Factor Retriever selects factors with higher potential by a trade-off among factors' intrinsic quality \& diversity, topological structure, and lineage. The DAG-aware Factor Generator utilizes a multi-agent structure and factors' topology to generate new factors.}
  \vspace{-1em}
  \label{fig:main}
\end{figure*}

\MYMETHOD~ is a closed-loop system designed to address this dual problem. It comprises a Bayesian Factor Retriever to solve the strategic retrieval task and a DAG-aware Factor Generator to perform targeted generation. The overall illustration of \MYMETHOD~ is in Figure \ref{fig:main}.

\subsection{Bayesian Factor Retriever}
To effectively select parent factors, we require a framework that can reason about the multi-level structural information inherent in the factor evolution process. Inspired by Bayesian modeling, we provide a principled way to integrate evidence from various sources—the factor's individual performance, its lineage and the topological structure it occupies, and its relationship to the broader pool.

The Retriever's task is to identify the parent factors with the greatest optimization potential. This selection problem is framed within a Bayesian framework to balance the trade-off between a factor's individual merit and its potential contribution to the collective factor pool with a topological structure $\mathcal{G}$. We seek to rank each factor $F \in \mathcal{F}$ by its probability of producing a high-value descendant $F_{new}$, given our current factor pool $\mathcal{D}$:
\begin{align}
\begin{split}
 \arg\max_{F \in \mathcal{F}} \mathbb{E}[ \text{Qual}(F_{new}) | \text{parent}(F_{new})=F, \mathcal{D}] \\
  \propto \frac{P(F_{new}) P(\mathcal{D} | F_{new})}{P(\mathcal{D})}\propto P(F_{new}) P(\mathcal{D} | F_{new})
 \end{split}
\end{align}
The prior term $P(F_{new})$  captures the intrinsic potential of a factor. The likelihood term $P(\mathcal{D} | F_{new})$ assesses the value a new descendant would add to the pool's ecosystem. The evidence term $ P(\mathcal{D})$ is a constant here.

\subsubsection{Prior: Trading Off Quality Against Over-Optimization Risk}

The prior probability $P(F_{new})$ is approximated by the quality of its parent $F$. However,$F$ with high performance alone can be misleading. A factor that is the result of a long optimization chain or has been frequently used for generation may yield diminishing returns. Our prior is therefore designed to explicitly trade off a factor's demonstrated quality against these structural risks, considering the topological structure of $F$ in $\mathcal{G}$.

We define the prior score for a candidate parent $F$ as:
\begin{align}
\begin{split}
\label{equ::prior}
    P(F) = & \underbrace{\sigma\left(\frac{\text{Qual}(F) - \mu_{Qual(\mathcal{F})}}{\varsigma_{Qual(\mathcal{F})}}\right)}_{\text{Normalized Quality}} \cdot \underbrace{(1 - \gamma)^{\text{depth}(F)}}_{\text{Depth Penalty}} \\
    & \cdot \underbrace{(1 - \omega)^{k(F)}}_{\text{Retrieval Penalty}}
\end{split}
\end{align}
Here, $\text{Qual}(F)$ denotes the factor’s risk-adjusted performance, normalized across the pool $\mathcal{F}$; $\sigma$ is a sigmoid mapping function; $\mu_{\text{Qual}(\mathcal{F})}$ and $\varsigma_{\text{Qual}(\mathcal{F})}$ are the mean and standard deviation of $\text{Qual}(\cdot)$ over $\mathcal{F}$, respectively; $\text{depth}(F)$ is the depth of factor $F$ in the graph $\mathcal{G}$; and $k(F)$ denotes the retrieval times of $F$ during the retrieval phase. This quality score is tempered by two topological penalties that represent the risk side of the trade-off:
\begin{itemize}[left=0.3cm,partopsep=0pt,topsep=-2pt,itemsep=-1pt]
    \item \textbf{Depth Penalty:} Addresses the trade-off between trusting a deeply optimized factor and the risk of it being overfitted, balanced by hyperparameter $\gamma$.
    \item \textbf{Retrieval Penalty:} Navigates the trade-off between exploiting a known successful parent and the need to explore other, less-mined areas of the search space, balanced by hyperparameter $\omega$.
\end{itemize}

\subsubsection{Likelihood: Assessing Contribution to the Collective}

The likelihood term $P(\mathcal{D} | F_{new})$ estimates the marginal utility of a new factor. It essentially assesses how much this new factor improves the collective pool's quality rather than just its own individual merit. Our estimation strategy adapts based on whether its parent factor $F$ has a proven history of generating successful descendants. We therefore distinguish between two types of factors in the DAG:
\begin{itemize}[left=0.3cm,partopsep=0pt,topsep=-2pt,itemsep=-1pt]
    \item \textbf{Leaf Factors:} Factors that have not yet produced any successful children. They represent unexplored optimization paths.
    \item \textbf{Non-Leaf Factors:} Factors that have successfully parented at least one child. They have a demonstrated history of generative potential.
\end{itemize}

\paragraph{For Leaf Factors} For a leaf factor, we have no past generation offspring. The core idea is to approximate its potential by holistically measuring its own novelty along with the whole factor pool $\mathcal{F}$, trading off among three distinct dimensions of diversity: value, semantics, and syntax. A factor that is novel across all these dimensions is more likely to spawn a truly unique descendant. Then $P(\mathcal{D} | F_{new}) $ can be estimated by:
\begin{align}
\begin{split}
   \text{ValDiv}(F, \mathcal{F}) \cdot \text{SemDiv}(F, \mathcal{F}) \cdot \text{SynDiv}(F, \mathcal{F})
\end{split}
\end{align}
These three components are:
\begin{itemize}
    \item \textbf{ValDiv (Value Diversity):} Measure the novelty of the factor's numerical output, implemented via its average Pearson correlation with other factors in $\mathcal{F}$:
    \begin{equation}
        1 - \|\frac{1}{|\mathcal{F}|} \sum_{f \in \mathcal{F}}Corr(F, f)\|
    \label{equ::valdiv}
    \end{equation}
    \item \textbf{SemDiv (Semantic Diversity):} Measure the novelty of the factor's underlying financial logic, implemented via the cosine similarity (CosSim) of the embedding of its LLM-generated explanation:
    \begin{equation}
    \label{equ::likelihood::embedding}
        \sigma(1 - \text{CosSim}(F, \mathcal{F}))
    \end{equation}
    
    \item \textbf{SynDiv (Syntactic Diversity):} Measure the novelty of the factor's mathematical structure, implemented via its normalized edit distance (ED) to other factors:
    \begin{equation} \frac{1}{|\mathcal{F}|} \sum_{f \in \mathcal{F}}
        \frac{ED(F, f)}{len(F) + len(f)}
    \end{equation}
\end{itemize}

\paragraph{For Non-Leaf Factors} For a non-leaf factor, we can make a more informed projection by extrapolating from its past success. The trade-off here is between parents that produce marginally better but similar children, versus those that spawn a wide array of diverse descendants.
\begin{align}
\begin{split}
\label{equ::nlf}
        P(\mathcal{D} | F_{new}) \approx \overline{\text{PG}}(\mathcal{C}(F))  \cdot \text{Spar}(\mathcal{C}(F))
\end{split}
\end{align}
Here, $\overline{\text{PG}}$ measures the average quality gain in percentage from the parent to its existing children. $\mathcal{C}(\cdot)$ denotes the children of the current factor. The $\text{Spar}$ term embodies a crucial trade-off on children's sparsity. We reward parents whose children are not only distinct from the parent (vertical diversity) but also distinct from each other (horizontal diversity). This encourages the discovery of parents that unlock multiple, independent optimization avenues. We formalize this as the product of two distinct sparsity measures:
\begin{equation}
    \text{Spar}(\mathcal{C}(F)) = \text{Spar}_{\text{p-c}}(F) \cdot \text{Spar}_{\text{c-c}}(F)
\end{equation}
    $\text{Spar}_{\text{p-c}}$ is the parent-child sparsity, measuring how much the children diverge from their parent:
\begin{equation}
    1 - \frac{1}{|\mathcal{C}(F)|} \sum_{f \in \mathcal{C}(F)} \operatorname{Corr}(F, f)
\end{equation}
    $\text{Spar}_{\text{c-c}}$ is the inter-child sparsity, measuring how much the children diverge from each other. For factors with a single child, we set $\text{Spar}_{\text{c-c}}(F)=1$. Otherwise, it is defined as:
\begin{equation}
     1 - \frac{1}{\binom{|\mathcal{C}(F)|}{2}}
    \sum_{\substack{f_i, f_j \in \mathcal{C}(F), i < j}}
    \text{Corr}(f_i, f_j)
\end{equation}
Finally, we calculate the total score for each factor. We rank leaf and non-leaf nodes based on their respective scores and select the global top-$k$ candidates to pass to the Factor Generator, thus completing the loop.

\subsection{DAG-aware Factor Generator}
Once the Retriever selects a parent factor $F_p$, the Generator's task is to produce novel and valid offspring. Rather than naively prompting an LLM for improvements—a process prone to syntactic errors and trivial variations—we structure generation as a three-stage, DAG-aware workflow that emulates a sequence of specialized analytical roles.

First, an Analyst agent leverages the parent's entire evolutionary path, its trace $T(F_p)$, to formulate a set of diverse and context-aware modification strategies, $\{\mathcal{S}\}$. This initial step transforms the historical optimization trajectory encoded in the DAG into actionable plans for future mutations, directly addressing the core of \MYMETHOD. This process is formalized as:
\begin{equation}
\label{equ::strategy}
    \{\mathcal{S}_1, \dots, \mathcal{S}_m\} = G_{\text{strategy}}(F_p, T(F_p))
\end{equation}
Next, an Execution agent translates each abstract strategy $\mathcal{S}_i$ into a concrete candidate factor expression, $F'_{c,i}$. This stage separates the high-level financial and structural reasoning from the precise task of expression synthesis:
\begin{equation}
\label{equ::generation}
    F'_{c,i} = G_{\text{synth}}(F_p, \mathcal{S}_i) \quad \forall i \in \{1, \dots, m\}
\end{equation}
Finally, a Validator agent subjects each candidate $F'_{c,i}$ to a rigorous check for syntactic correctness and adherence to predefined constraints, filtering out invalid expressions. This entire workflow constitutes the generation function $G$. The set of newly validated factors $\{F_c\} \subseteq \{F'_c\}$ is then added to the DAG with new edges from $F_p$, completing the evolutionary loop and expanding the graph for the next retrieval cycle. The detailed design of Factor Generator is provided in the Appendix ~\ref{appendix::detail:generator}.
\subsection{Dynamic Factor Integrator}
\label{sec::integrator}
We follow the approach proposed by AlphaForge ~\cite{shi2025alphaforge} to dynamically re-select recently effective factors and integrate them into a ``Mega'' factor $F_{\hat{y}}$ for portfolio construction.

%% file: sections/Evaluation.tex
\FloatBarrier
\begin{table*}[t]
    \centering
    \caption{\textbf{Performance Comparison} on the CSI 300, CSI 500 and CSI 1000.
    \emph{Bold} and \underline{underlined} numbers denote the best and second-best results.
    $^{\uparrow}$/$^{\downarrow}$ indicate higher/lower is better.}
    \label{tab:performance_comparison}

    % spacing tweaks (different "team style")
    \setlength{\tabcolsep}{6pt}
    \renewcommand{\arraystretch}{1.118}
    \small

    % subtle color theme change (from gray to light blue)
    \definecolor{myhi}{RGB}{230,242,255}

    % ---------- Panel title (NO caption, so no "Table 2" / "(a)") ----------
    \vspace{0.2em}
    {\centering \textbf{Predictive Power (\%)}\par}
    \vspace{0.2em}

    \scalebox{0.89}{
    \begin{tabular}{l|cccc|cccc|cccc}
        \hline\hline
        \multirow{2}{*}{\textbf{Method}} &
        \multicolumn{4}{c|}{\textbf{CSI300}} &
        \multicolumn{4}{c|}{\textbf{CSI500}} &
        \multicolumn{4}{c}{\textbf{CSI1000}} \\
        \cline{2-13}
        & \textit{IC}$^{\uparrow}$ & \textit{ICIR}$^{\uparrow}$ & \textit{RIC}$^{\uparrow}$ & \textit{RICIR}$^{\uparrow}$
        & \textit{IC}$^{\uparrow}$ & \textit{ICIR}$^{\uparrow}$ & \textit{RIC}$^{\uparrow}$ & \textit{RICIR}$^{\uparrow}$
        & \textit{IC}$^{\uparrow}$ & \textit{ICIR}$^{\uparrow}$ & \textit{RIC}$^{\uparrow}$ & \textit{RICIR}$^{\uparrow}$ \\
        \hline
        Alpha158   & 3.91 & 25.80 & 5.77 & 37.61 & 5.24 & 38.16 & 7.72 & 54.78 & 7.85 & 51.42 & 10.10 & 65.85 \\
        GP         & 1.36 & 9.97  & 1.96 & 14.63 & 2.83 & 20.76 & 5.82 & 41.67 & 4.85 & 46.77 & 6.98 & 69.37 \\
        AlphaGen   & 4.93 & 29.41 & \underline{6.32} & 38.95 & 4.77 & 38.13 & 6.35 & 49.54 & 7.44 & 54.85 & 9.68 & 71.99 \\
        AlphaForge & 4.56 & 29.16 & 5.17 & 33.14 & 5.17 & 31.63 & 7.85 & 49.90 & 8.24 & 57.84 & 10.30 & 68.90 \\
        AlphaQCM   & 4.03 & 32.99 & 4.53 & 36.13 & 5.06 & 34.17 & 7.73 & 52.09 & 7.90 & 56.31 & 10.63 & \underline{76.79} \\
        AlphaSAGE  & \underline{5.02} & \underline{35.82} & 6.31 & \underline{44.58} & 4.46 & 35.62 & 6.58 & 54.89 & 7.27 & 58.45 & 9.41 & 76.77 \\
        AlphaAgent & 4.27 & 25.66 & 5.45 & 30.94 & 5.50 & \underline{39.90} & 7.45 & \underline{54.90} & 8.01 & 54.00 & 10.95 & 60.86 \\
        R\&D-Agent(Q) & 4.88 & 29.39 & 6.30 & 36.73 & \underline{5.81} & 37.72 & \underline{7.94} & 52.35 & \underline{8.58} & \underline{61.00} & \underline{11.25} & 76.03 \\
        \rowcolor{myhi}
        \textbf{\MYMETHOD} & \textbf{5.84} & \textbf{39.02} & \textbf{7.20} & \textbf{46.94}
                           & \textbf{6.26} & \textbf{52.39} & \textbf{8.78} & \textbf{73.18}
                           & \textbf{9.04} & \textbf{70.49} & \textbf{11.35} & \textbf{88.02} \\
        \hline\hline
    \end{tabular}
    }

    % ---------- Panel title (NO caption, so no "Table 3" / "(b)") ----------
    \vspace{0.8em}
    {\centering \textbf{Portfolio Construction}\par}
    \vspace{0.2em}

    \scalebox{0.92}{
    \begin{tabular}{l|ccc|ccc|ccc}
        \hline\hline
        \multirow{2}{*}{\textbf{Method}} &
        \multicolumn{3}{c|}{\textbf{CSI300}} &
        \multicolumn{3}{c|}{\textbf{CSI500}} &
        \multicolumn{3}{c}{\textbf{CSI1000}} \\
        \cline{2-10}
        & \textit{AR}$^{\uparrow}$ & \textit{MDD}$^{\downarrow}$ & \textit{SR}$^{\uparrow}$
        & \textit{AR}$^{\uparrow}$ & \textit{MDD}$^{\downarrow}$ & \textit{SR}$^{\uparrow}$
        & \textit{AR}$^{\uparrow}$ & \textit{MDD}$^{\downarrow}$ & \textit{SR}$^{\uparrow}$ \\
        \hline
        Alpha158   & 6.03\%  & 25.29\% & 0.3925 & 10.39\% & 25.78\% & 0.5682 & 9.32\%  & 34.10\% & 0.4372 \\
        GP         & 3.46\%  & 39.06\% & 0.1882 & 6.52\%  & \underline{23.46\%} & 0.3421 & 10.23\% & 38.16\% & 0.4078 \\
        AlphaGen   & \underline{6.19\%} & 31.40\% & 0.3425 & 7.49\%  & 31.85\% & 0.3429 & 10.96\% & 36.12\% & 0.4416 \\
        AlphaForge & 1.01\%  & 35.34\% & 0.0694 & 8.09\%  & 23.09\% & 0.4257 & 13.72\% & \textbf{31.86\%} & 0.5945 \\
        AlphaQCM   & 3.42\%  & \underline{24.13\%} & 0.1973 & 8.86\%  & 30.13\% & 0.4118 & 13.98\% & 32.69\% & 0.5786 \\
        AlphaSAGE  & 4.15\%  & 30.79\% & 0.1832 & 10.46\% & 23.48\% & \underline{0.5220} & \underline{14.98\%} & 32.03\% & \underline{0.6099} \\
        AlphaAgent & 3.43\%  & 31.74\% & 0.1883 & 6.91\%  & 31.57\% & 0.3588 & 12.29\% & 32.18\% & 0.5861 \\
        R\&D-Agent(Q) & 6.06\% & 31.38\% & \underline{0.3456} & \underline{12.36\%} & 30.74\% & 0.5165 & 13.38\% & 33.53\% & 0.5968 \\
        \rowcolor{myhi}
        \textbf{\MYMETHOD} & \textbf{7.50\%} & \textbf{22.25\%} & \textbf{0.4411}
                           & \textbf{17.45\%} & \textbf{22.98\%} & \textbf{0.8262}
                           & \textbf{16.68\%} & \underline{31.95\%} & \textbf{0.6475} \\
        \hline\hline
    \end{tabular}
    }

    \vspace{-0.8em}
\end{table*}

\section{Empirical Experiments}

\subsection{Experiment Setup}

\textbf{Evaluation metrics: } Following prior work \cite{tang2025alphaagent, chen2025alphasage}, we employed two types of metrics for model evaluation.(1): \textbf{Predictive Power}, including Information Coefficient (IC), IC Information Ratio (ICIR), Rank Information Coefficient
(RIC), RIC Information Ratio (RICIR). These metrics are all computed against the 20-day forward returns.
(2): \textbf{Portfolio Construction}, including Annualized Return (AR),
Maximum Drawdown (MDD), Sharpe Ratio (SR). 
Definitions about evaluation metrics and backtest settings are provided in the Appendix ~\ref{appendix::detail:metrics}.

\textbf{Datasets:} Following ~\cite{tang2025alphaagent, chen2025alphasage}, we conduct experiments on the three most popular Chinese stock pools: the CSI 300 (large cap), the CSI 500 (mid cap), and the CSI 1000 (small cap). The split of the training/validation/test set is defined as 2010-01-01 to 2020-12-31 / 2021-01-01 to 2022-06-30 / 2022-07-01 to 2025-06-30.

\textbf{Baselines: } We compare \MYMETHOD~ with various baselines: (1) Expert-Collected factor pool: Alpha158 ~\cite{qlib} (2) DFG methods: AlphaGen ~\cite{yu2023alphagen}, AlphaForge ~\cite{shi2025alphaforge}, AlphaQCM ~\cite{alphaqcm}, AlphaSAGE ~\cite{chen2025alphasage}. (3) IFE methods: GP ~\cite{chen2021gp}, AlphaAgent ~\cite{tang2025alphaagent}, R\&D-Agent(Q) ~\cite{li2025rdagentquant}. More details about baselines and implementation are in the Appendix ~\ref{appendix::detail:baseline}.

\textbf{Implementation details}: We use Deepseek V3.1~\cite{deepseekai2024deepseekv3technicalreport} as the backbone LLM for \MYMETHOD~and two LLM-based baselines. We use Qwen 3 Embedding-4B~\cite{qwen3embedding} as the embedding model in the Equation~\ref{equ::likelihood::embedding}.  We use the absolute value of the factor's ICIR during the training period as the quality measurement in the Equation ~\ref{equ::prior}. The number of generated factors in Equation  ~\ref{equ::strategy},\ref{equ::generation} is set to 5. Following ~\cite{yu2023alphagen, shi2025alphaforge, chen2025alphasage}, we set the pool capacity to 50 and the length threshold for each factor to 40 for~\MYMETHOD~and all baselines. Following ~\cite{yu2023alphagen}, when the factor pool hits the size limit, we filter out the factor with the lowest quality, while the filtered factors are still maintained in the $\mathcal{G}$ to ensure the completeness of factor topology. The depth penalty $\gamma$ and the retrieval penalty $\omega$ are set to $0.05$ and $0.10$. All baselines use the same dynamic factor integrator in Section~\ref{sec::integrator} for fair comparison.

\subsection{Main Experiments}
Table~\ref{tab:performance_comparison} summarizes the empirical results across three distinct datasets, from which two primary conclusions can be drawn. First, \MYMETHOD~ exhibits superior predictive power for future stock returns, consistently achieving the highest IC, RIC, and AR.  Recent research like ~\cite{tang2025alphaagent, li2025rdagentquant} also exhibits promising results, underscoring the significant potential of leveraging  LLMs for alpha factor discovery. Second, \MYMETHOD~ demonstrates enhanced stability against market regime shifts. This is substantiated by its significantly higher ICIR, RICIR, and SR, coupled with a lower MDD.

\subsection{Backtest Experiments}

% and the top-k candidates for retrieval is set to 10.
\begin{figure}[htbp!]
    \centering
    \vspace{-1em}
    \includegraphics[width=0.82\linewidth]{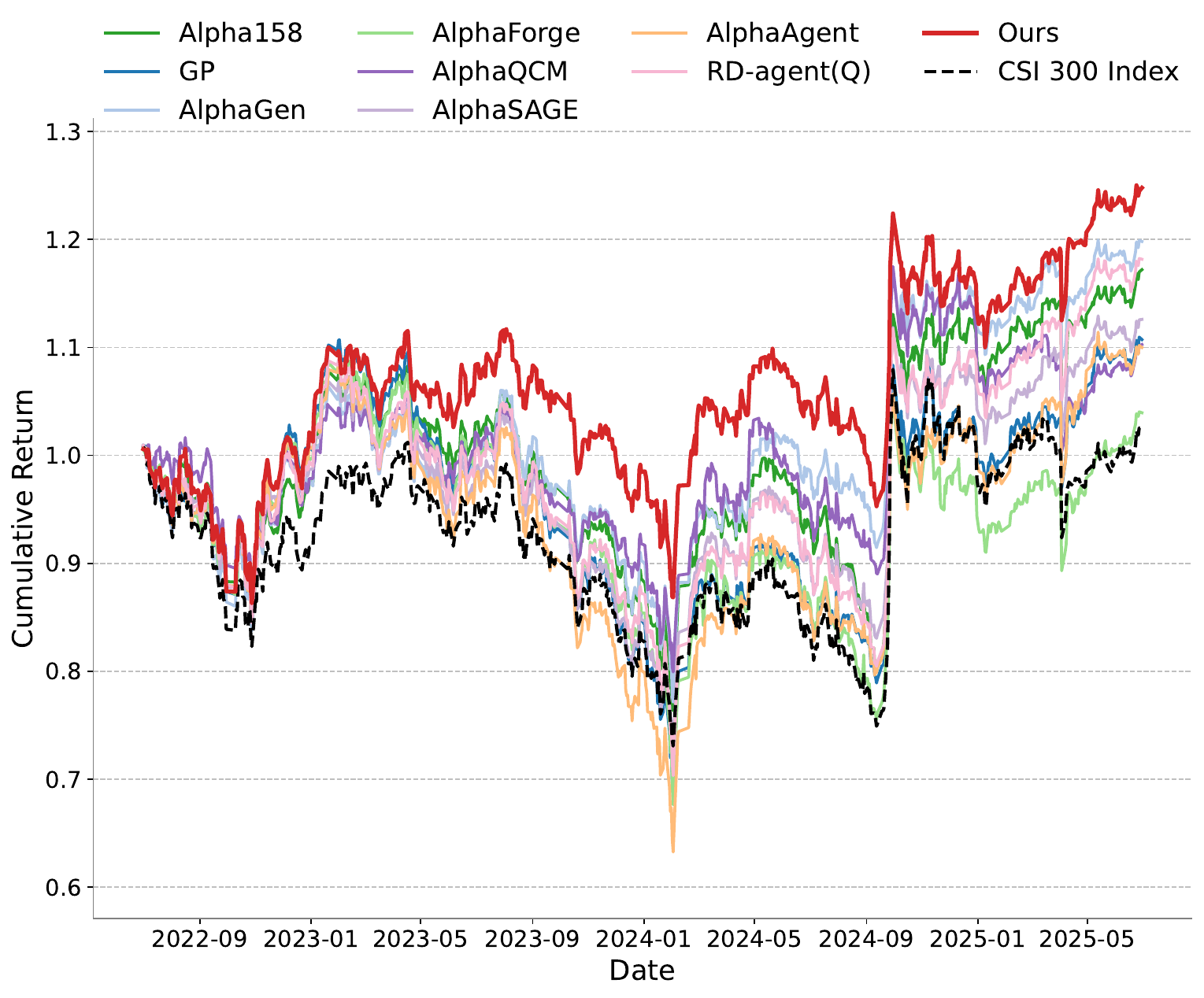}
    \caption{Backtest curve on the CSI 300.}
    \vspace{-1em}
    \label{fig:csi_300_backtest}
\end{figure}

To validate the practical efficacy of \MYMETHOD~in realistic trading environments, we conducted comprehensive backtesting experiments across all three distinct datasets. As illustrated by the results on the CSI 300 in Figure~\ref{fig:csi_300_backtest}, ~\MYMETHOD~ not only maintains a leading edge in cumulative returns throughout the vast majority of the backtest period but also demonstrates superior resilience. Specifically, it exhibited more controlled drawdowns and faster recoveries during significant market stress events, such as the bear market from late 2023 to early 2024 and the tariff-induced turmoil in April 2025. For completeness, backtesting results for the CSI 500 and the CSI 1000, along with details of the experimental setup, are deferred to the Appendix~\ref{appendix::detail:backtest}.
\begin{table*}[t]
\centering
\caption{Ablation study of ~\MYMETHOD's core components on three datasets. We report the IC, ICIR, RIC, and RICIR (\%) metrics. The results demonstrate the effectiveness of the Bayesian Factor Retriever and the DAG-aware Factor Generator.}
\label{tab:ablation}

\definecolor{myhi}{RGB}{230,242,255}
\setlength{\tabcolsep}{7pt}
\renewcommand{\arraystretch}{1.06}
\small

\resizebox{\textwidth}{!}{%
\begin{tabular}{l|cccc|cccc|cccc}
\hline\hline
\multirow{2}{*}{\textbf{Method}}
& \multicolumn{4}{c|}{\textbf{CSI 300}}
& \multicolumn{4}{c|}{\textbf{CSI 500}}
& \multicolumn{4}{c}{\textbf{CSI 1000}} \\
\cline{2-13}
& IC & ICIR & RIC & RICIR
& IC & ICIR & RIC & RICIR
& IC & ICIR & RIC & RICIR \\
\hline

% ---- FIX: keep vertical rules continuous by splitting into 1|4|4|4 blocks ----
\multicolumn{1}{l|}{\textit{\textbf{Retriever Ablation}}}
& \multicolumn{4}{c|}{} & \multicolumn{4}{c|}{} & \multicolumn{4}{c}{} \\
\quad w/ Random Retriever
& 2.95 & 18.71 & 3.18 & 21.65
& 3.82 & 21.44 & 4.17 & 26.93
& 5.64 & 32.45 & 6.88 & 43.30 \\
\quad w/ Heuristic Retriever
& 4.54 & 35.15 & 5.92 & 35.71
& 5.11 & 41.86 & 6.08 & 39.03
& 7.74 & 50.09 & 8.15 & 58.33 \\
\quad w/ MCTS Retriever
& 4.75 & 35.20 & 5.94 & 36.86
& 5.09 & 43.25 & 6.11 & 38.71
& 7.85 & 49.94 & 8.37 & 59.27 \\
\quad w/o Prior
& 4.13 & 34.88 & 5.68 & 34.42
& 4.82 & 40.07 & 5.99 & 36.87
& 7.60 & 47.72 & 7.95 & 58.01 \\
\quad w/o Topology penalty
& 5.06 & 36.37 & 6.27 & 40.05
& 5.80 & 44.38 & 8.27 & 69.24
& 8.37 & 54.62 & 9.68 & 64.89 \\
\quad w/o Likelihood
& 4.09 & 34.99 & 5.66 & 34.39
& 4.86 & 40.02 & 6.01 & 36.98
& 7.49 & 46.96 & 7.67 & 57.70 \\
\quad w/o NLF
& 5.15 & 37.01 & 6.40 & 42.17
& 5.93 & 43.88 & 8.29 & 68.94
& 8.45 & 56.08 & 10.13 & 67.96 \\
\hline

% ---- FIX here as well ----
\multicolumn{1}{l|}{\textit{\textbf{Generator Ablation}}}
& \multicolumn{4}{c|}{} & \multicolumn{4}{c|}{} & \multicolumn{4}{c}{} \\
\quad w/ CoT Generator
& 5.11 & 31.79 & 6.08 & 39.17
& 5.41 & 39.70 & 8.35 & 63.00
& 8.03 & 60.12 & 9.75 & 66.54 \\
\hline

\rowcolor{myhi}
\textbf{\MYMETHOD}
& \textbf{5.84} & \textbf{39.02} & \textbf{7.20} & \textbf{46.94}
& \textbf{6.26} & \textbf{52.39} & \textbf{8.78} & \textbf{73.18}
& \textbf{9.04} & \textbf{70.49} & \textbf{11.35} & \textbf{88.02} \\
\hline\hline
\end{tabular}
}
\vspace{-0.8em}
\end{table*}

\subsection{Ablation Study}
Table~\ref{tab:ablation} summarizes the ablation study, evaluating components across the retriever and generator modules:

\begin{itemize}[left=0.3cm,partopsep=-2pt,topsep=-2pt,itemsep=2pt]

    \item \textbf{w/ Random Retriever}: Replacing our retriever with random selection leads to a sharp performance drop, proving that ignoring factor relationships results in noisy optimizations.

    \item \textbf{w/ Heuristic Retriever}: Ranking factors by individual metrics~\cite{alphaeval} without structural context yields sub-optimal results, proving the necessity of strategic information encoded in the evolutionary graph.

    \item \textbf{w/ MCTS Retriever}: Local-view MCTS~\cite{shi2025navigating} falls short of \MYMETHOD, confirming a global topological view is superior to local lineage tracking.

    \item \textbf{w/o Prior / Topology Penalty}: Removing the prior term or the topology penalty degrades performance, demonstrating that both intrinsic factor quality and graph positioning are essential for effective retrieval.

    \item \textbf{w/o Likelihood / NLF}: The decline in performance upon removing these components indicates that pool-level context and ancestral lineage provide critical, non-redundant information for discovery.

    \item \textbf{w/ CoT Generator}: Substituting the DAG-aware generator with simple alpha chains confirms that the DAG structure encodes richer interactions and better guides the evolutionary search.

\end{itemize}

\subsection{Parameter Sensitivity Study}
\begin{figure}[ht!]
    \centering
    \includegraphics[width=0.92\linewidth]{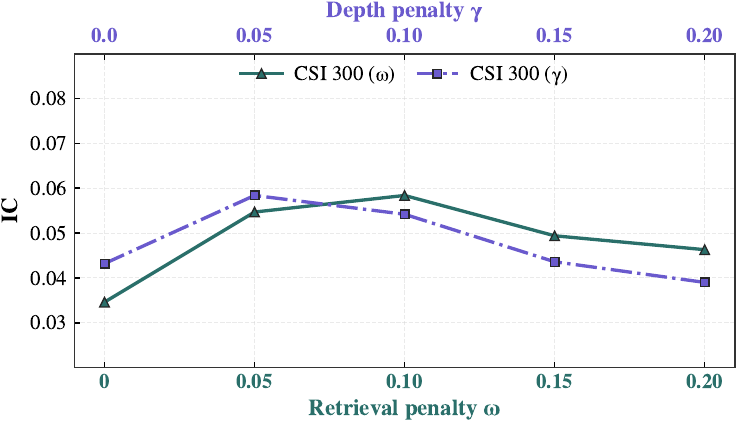}
    \caption{Sensitivity study of~\MYMETHOD.}
    \label{fig:parameter_senivity}
    \vspace{-1em}
\end{figure}

We investigate the sensitivity of \MYMETHOD~ to the hyperparameters $\gamma$ and $\omega$ in Equation~\ref{equ::prior} on the CSI 300 dataset. As shown in Figure~\ref{fig:parameter_senivity}, \MYMETHOD~ demonstrates considerable robustness, with performance remaining stable across a wide range of values for both parameters, especially within the 0.05 to 0.15 interval.

Extreme parameter values degrade performance by disrupting the balance between exploitation and exploration. An overly small $\gamma$ or $\omega$ leads to over-exploitation, risking the selection of over-fitted factors or repeatedly using the same parents. Conversely, excessively large values cause under-exploitation, with a high $\gamma$ prematurely penalizing deep, promising factors and a high $\omega$ forcing exploration away from high-potential areas.

\subsection{Visualization}
\begin{figure}[htbp]
    \centering
    \includegraphics[width=0.92\linewidth]{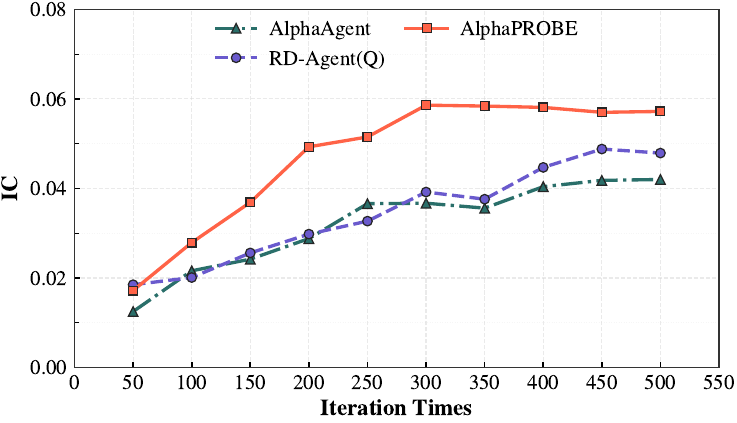}
    \caption{IC dynamics of \MYMETHOD~and two LLM-based methods on the CSI 300  test set over training iterations. For fair comparison, one iteration here means the backbone LLM generate a new factor to be evaluated. }
    \label{fig:efficiency}
\end{figure}
To illustrate the factor discovery process, we present a  visualization on factors for interday price movements, with the resulting DAG shown in Figure~\ref{fig:case_study}. The process can originate from a simple, human-designed seed factor such as \texttt{Div(Sub(Less(\$open, \$close), \$low), \$open)} about interday price movements. Our framework identifies this as a promising parent and generates a more complex, high-quality descendant, \texttt{TsCorr(Sub(\$close, \$open), Sub(\$low, TsMin(\$low, 5)), 5)}, by introducing a time-series correlation. Besides, we can observe that \MYMETHOD~ can grab different potential of candidate factors. This evolutionary step highlights how \MYMETHOD~ establishes clear optimization pathways. Further details on all factors in Figure~\ref{fig:case_study} are in Table~\ref{table:factors} the Appendix.

\subsection{Efficiency Analysis}
We further investigate the training efficiency of \MYMETHOD~by comparing its convergence speed with LLM-based baselines. As illustrated in Figure~\ref{fig:efficiency}, by leveraging global information with structural topology, the retriever in \MYMETHOD~identifies the most promising factors, enabling it to achieve better results with fewer training iterations and thus demonstrating its efficiency.
% To investigate the diversity of the factor pool $\mathcal{F}$ mining by \MYMETHOD~, we conduct a visualization case study on the CSI 300 dataset using t-SNE ~\cite{tsne} during the test period. 

% Figure ~\ref{fig:visual} visualizes the factor set $\mathcal{F}$ discovered by \MYMETHOD~in a 2D latent representation space. Each point corresponds to a factor, colored by its average mean absolute correlation with all other factors in the set. The observation of visualization is twofold: first, the factors are widely dispersed across the embedding space, indicating significant diversity in their latent representations. Second, the vast majority of factors exhibit low inter-correlation, as signified by the color scale on the right. This empirically confirms that \MYMETHOD~ effectively mines a diversified portfolio of alpha factors.

%% file: sections/Conclusion.tex
\section{Conclusion}
In this paper, we introduce ~\MYMETHOD, a closed-loop retriever \& generator framework designed for alpha mining. \MYMETHOD~ leverages a Bayesian Factor Retriever to capture factors' exploitation-exploration trade-off beyond factor topology, and a DAG-aware Factor Generator to perform context-aware optimization. This approach offers enhanced predictive performance, stability, and hierarchy on three Chinese stock pools with distinct styles.

Looking ahead, we anticipate \MYMETHOD~will extend beyond traditional daily-frequency scenarios to high-frequency and fundamental factor mining.

%% file: sections/Appendix.tex
\newpage
\onecolumn

\section{Appendix}
\setcounter{equation}{25}
\subsection{Experiment details}
\subsubsection{Details about evaluation metrics}
\label{appendix::detail:metrics}
In this section, we define the metrics used to evaluate the performance of a predictive model (or alpha factor), denoted by $\alpha$. On any given day $d$, the model processes a set of features for all assets in the universe, $X_d$, to generate a vector of predictive scores, $\alpha(X_d)$. These scores are evaluated against the vector of actual next-period outcomes, $y_d$ (e.g., forward returns). All statistical moments, such as covariance ($\operatorname{Cov}$) and variance ($\operatorname{Var}$), are computed cross-sectionally across the asset universe for that specific day.

The foundational metric is the daily cross-sectional Pearson correlation between predictions and outcomes:
\begin{equation*}
    \rho_d = \frac{\operatorname{Cov}(\alpha(X_d), y_d)}{\sqrt{\operatorname{Var}(\alpha(X_d)) \operatorname{Var}(y_d)}}.
\end{equation*}
For metrics based on portfolio simulation, let $R_d$ denote the portfolio return constructed from the factor scores $\alpha(X_d)$ on day $d$. Let $K$ be the number of periods per year (e.g., $K=252$ for daily data), $r_{f,d}$ be the risk-free rate, and $W_t$ be the cumulative wealth at time $t$:
\begin{equation*}
    W_t = \prod_{u=1}^{t} (1 + R_u).
    % Note: I changed the original \sum to \prod for cumulative wealth, which is standard.
    % If the original's \sum was intentional for a specific arithmetic return model, it can be reverted.
\end{equation*}

\paragraph{Information Coefficient (IC)}
The Information Coefficient is the absolute value of the time-series average of the daily cross-sectional correlations: $\text{IC} = |\mathrm{E}_d[\rho_d]|$.
It measures the cross-sectional predictive power of the factor—how well its predictions $\alpha(X_d)$ align with outcomes $y_d$. Using the absolute value isolates the magnitude of the correlation from its sign (as a consistently negative correlation is also predictive).

\paragraph{Information Ratio of IC (ICIR)}
\begin{equation*}
    \text{ICIR} = \frac{\mathrm{E}_d[\rho_d]}{\sqrt{\mathrm{Var}_d(\rho_d)}}.
\end{equation*}
Time-series consistency of cross-sectional predictability; the mean of daily correlations relative to its standard deviation. ICIR approximates a signal-to-noise measure (akin to a t-statistic for $\mathrm{E}[\rho_d]$), favoring factors that perform consistently over time.

\paragraph{Rank Information Coefficient (RIC)}
\begin{align*}
    \text{RIC} = \left|\mathrm{E}_d[\rho_d^{\text{rank}}]\right|, \qquad \rho_d^{\text{rank}} = \frac{\operatorname{Cov}(\operatorname{rank}(\alpha(X_d)), \operatorname{rank}(y_d))}{\sqrt{\operatorname{Var}(\operatorname{rank}(\alpha(X_d))) \operatorname{Var}(\operatorname{rank}(y_d))}}.
\end{align*}
 The Spearman correlation counterpart to IC, which evaluates if higher-ranked predictions correspond to higher-ranked outcomes. RankIC is robust to outliers and non-linear monotonic relationships.

\paragraph{Information Ratio of RankIC (RICIR)}
\begin{equation}
    \text{RICIR} = \frac{\mathrm{E}_d[\rho_d^{\text{rank}}]}{\sqrt{\mathrm{Var}_d(\rho_d^{\text{rank}})}}.
    \tag{30}
\end{equation}
Measure the time-series stability of the rank-based predictive power, prioritizing factors whose cross-sectional ordering of assets remains reliable over time.

\paragraph{Annualized Return (AR)}
\begin{equation*}
    \text{AR} = K \cdot \mathrm{E}_d[R_d].
\end{equation*}
The average economic value produced by the portfolio strategy induced by $\alpha$. When compounding is significant, geometric annualization is often preferred.

\paragraph{Maximum Drawdown (MDD)}
\begin{equation*}
    \text{MDD} = \max_t \left( 1 - \frac{W_t}{\max_{u \le t} W_u} \right).
\end{equation*}
 The largest peak-to-trough loss in cumulative wealth. It is a critical tail-risk metric not captured by variance alone, relevant for risk management and assessing investor experience.

\paragraph{Sharpe Ratio (SR) ~\cite{sharpe1998sharpe} (annualized, excess over risk-free)}
\begin{equation*}
    \text{SR} = \frac{\sqrt{K} \, \mathrm{E}_d[R_d - r_{f,d}]}{\sqrt{\mathrm{Var}_d(R_d - r_{f,d})}}.
\end{equation*}
The risk-adjusted return per unit of volatility for the $\alpha$-induced portfolio, enabling fair comparison across different strategies, asset universes, and rebalancing frequencies.

\subsubsection{Details about baselines}
\label{appendix::detail:baseline}
We conduct experiments on various competitive baselines. They can be separated into three categories:(1) Factors collected by human experts. (2) Decoupled Factor Generation(DFG). (3) Iterative Factor Evolution(IFE).

\paragraph{Expert-collected Factors}:

Alpha158~\cite{qlib} is a well-established factor pool constructed by Microsoft Qlib ~\footnote{\url{https://github.com/microsoft/qlib}} platform.

\paragraph{DFG methods}:
\vspace{-1em}
\begin{itemize}
    \item AlphaGen ~\cite{yu2023alphagen}: AlphaGen leverages the strong exploratory capabilities of
reinforcement learning (RL) to better explore the vast search space based on policy gradient algorithms. We use the open-source link ~\footnote{\url{https://github.com/RL-MLDM/alphagen}} to implement AlphaGen.
of formulaic alphas.
    \item AlphaForge ~\cite{shi2025alphaforge}: AlphaForge is a novel deep learning framework designed for formulaic alpha mining by a surrogate model to predict fitness score and a composite model to dynamically combine factors during test time combination. We use the open-source link ~\footnote{\url{https://github.com/DulyHao/AlphaForge}} to implement AlphaForge.
    \item AlphaQCM ~\cite{alphaqcm}: AlphaQCM is a powerful distributional reinforcement learning framework for alpha discovery relying on the unbiased estimation of variance derived from potentially biased quantiles. We use the open-source link ~\footnote{\url{https://github.com/ZhuZhouFan/AlphaQCM}} to implement AlphaQCM.
    \item AlphaSAGE ~\cite{chen2025alphasage}: AlphaSAGE is a recently proposed framework designed for alpha mining by unifying a GNN encoder for symbolic expressions, a GFlowNet generator that explores multiple high-reward modes, and a multi-signal training objective coupling predictive quality.We use the open-source link ~\footnote{\url{https://github.com/BerkinChen/AlphaSAGE}} to implement AlphaSAGE.
\end{itemize}
\paragraph{IFE methods}:
\begin{itemize}
    \item GP ~\cite{chen2021gp}: Genetic programming performs symbolic regression by evolving
expression trees via mutation and crossover, yielding human-readable formulas. We use the gplearn package to implement this baseline~\footnote{\url{https://gplearn.readthedocs.io/en/stable}}.
    \item AlphaAgent ~\cite{tang2025alphaagent}: AlphaAgent is a novel LLM-driven alpha mining framework with three key regularization mechanisms: originality enforcement,complexity control, and hypothesis alignment. We use the open-source link~\footnote{\url{https://github.com/RndmVariableQ/AlphaAgent}} to implement AlphaAgent.
    \item R\&D-Agent(Q) ~\cite{li2025rdagentquant}:R\&D-Agent(Quant) is a powerful LLM-driven framework for collaborative factor-model development in quantitative finance.We use the open-source link~\footnote{\url{https://github.com/microsoft/RD-Agent}} to implement R\&D-Agent(Q).
\end{itemize}

\subsubsection{More details about backtest experiments}
\begin{figure*}
    \centering
        \includegraphics[width=0.48\linewidth]{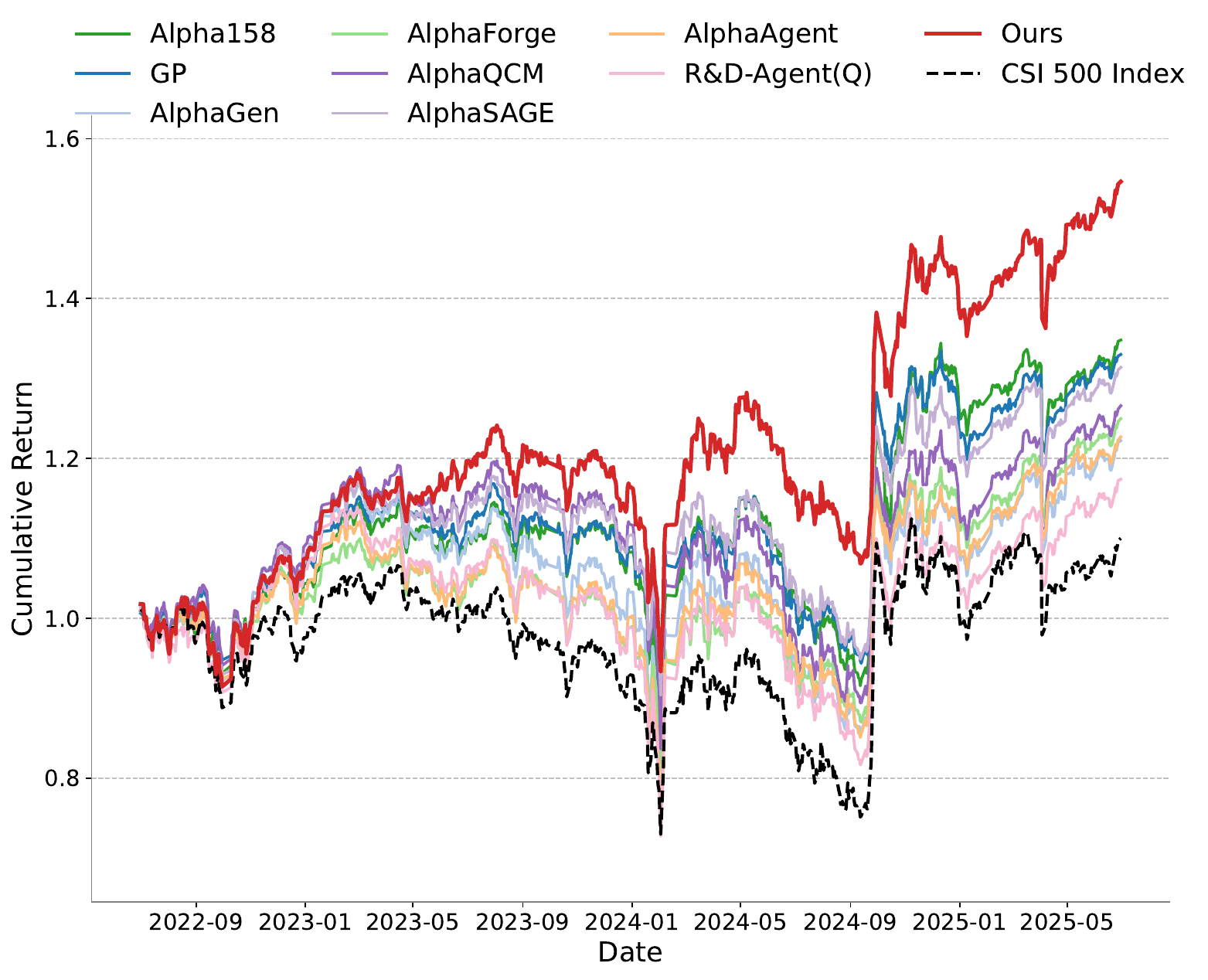}
        \caption{Backtest on the CSI 500 dataset}
        \label{fig:sub1}
\end{figure*}

\begin{figure*}
    \centering
        \includegraphics[width=0.48\linewidth]{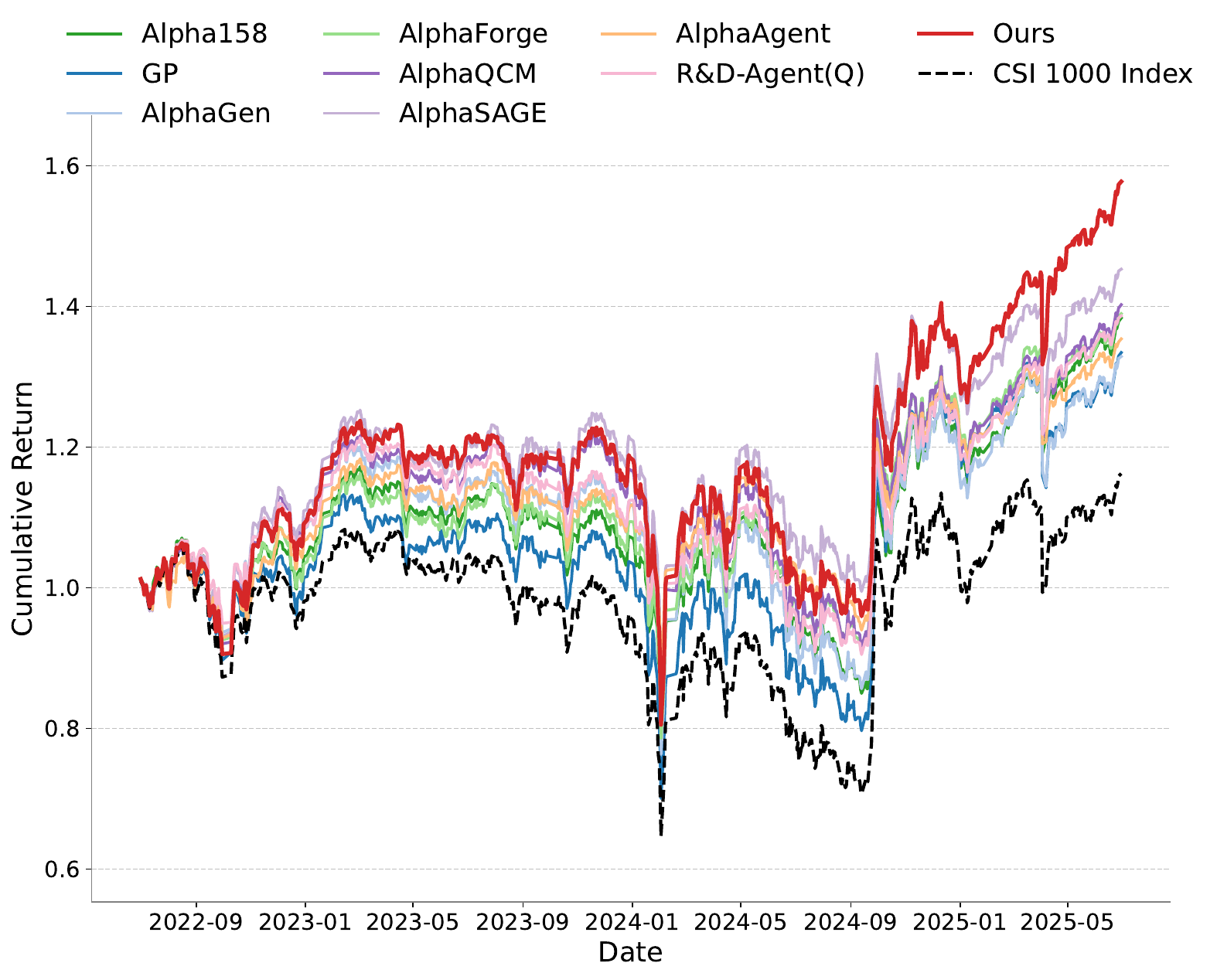}
        \caption{Backtest on the CSI 1000 dataset.}
        \label{fig:sub2}
\end{figure*}

During backtesting, we purchase the top 20\% of stocks each trading day and sell them after 20 days (long positions only), with a round-trip transaction
cost of 0.1\%. Figure ~\ref{fig:sub1} and ~\ref{fig:sub2} report the backtest results for \MYMETHOD~and all baselines on the CSI 500 and the CSI 1000 dataset.
\label{appendix::detail:backtest}
\subsection{More details about \MYMETHOD}

% \subsubsection{Definition of ED between factors}
% \label{appendix::detailed}

\subsubsection{Implementation of DAG-aware Factor Generator}
\paragraph{Strategy \& Execution Agent}
\label{appendix::detail:generator}

\begin{tcolorbox}[
  enhanced,                   % 开启增强功能
  breakable,                  % 支持跨页／跨栏
  colback=gray!5,             % 背景色
  colframe=blue!60!black,     % 边框色
  boxrule=0.8pt,              % 边框粗细
  arc=2mm,                    % 圆角半径
  left=4pt, right=4pt,        % 内边距
  top=4pt, bottom=4pt,
  width=\linewidth,           % 当前栏宽
  title={\bfseries System \& User Prompts},
  fonttitle=\bfseries
]
% 切换到等宽小号字体并启用两端对齐
{\ttfamily\small\justifying
\textbf{System prompts}
You are an expert on quantitative finance and alpha factor mining. Strictly follow the instructions given by user below. Make sure the output ONLY CONTAINS A  JSON FORMAT. Do not output anything else other than the JSON object, including code block markers like ``` or ```json.

\textbf{User prompts for features and operators definition}

The available features, constants and operators are listed below.

1. You can use the following features: 

   \$open, \$high, \$low, \$close: Opening, daily highest, daily lowest, and closing prices.
   
   \$vwap: Daily average price, weighted by the volume of trades at each price.
   
   \$volume: Trading number of shares.
   
2. You can use int constants eg:1, 3, 5,10, 20... etc during rolling(time-series) calculations, and float : 0.0001 0.01, 0.0, 1.0, 2.0 during arithmetic calculations. Other constants are not allowed.

3. The following operators are available:

(BEGIN OF FEATURES AND OPERATORS  DEFINITIONS)

    Abs(x): Absolute value of x
    
    Log(x): Natural logarithm of x
    
    SLog1p(x): Signed log transform: sign(input) times log of (1 plus the absolute value)
    
    Sign(x): Sign of x: 1 if x > 0, -1 if x < 0, 0 if x = 0
    
    Rank(x): Cross-sectional rank of x
    
    Add(x,y): x + y
    
    Sub(x, y): x - y
    
    \text{Mul(x, y): x * y}
    
    Div(x, y): x / y
    
    Pow(x, y): x raised to the power of y (x ** y) y must be a constant
    
    Greater(x, y): 1 if x > y, else 0
    
    Less(x, y): 1 if x < y, else 0
    
    GetGreater(x, y): x if x > y, else y
    
    GetLess(x, y): x if x < y, else y
    
    Ref(x, d): Value of x d days ago
    
    TsMean(x, d):  Rolling mean of x over the past d days
    
    TsSum(x, d): Rolling sum of x over the past d days
    
    TsStd(x, d): Rolling standard deviation of x over the past d days
    
    TsMin(x, d): Rolling minimum of x over the past d days
    
    TsMax(x, d): Rolling maximum of x over the past d days
    
    TsMinMaxDiff(x, d): Difference between TsMax(x, d) and TsMin(x, d)
    
    TsMaxDiff(x, d): Difference between current x and TsMax(x, d)
    
    TsMinDiff(x, d): Difference between current x and TsMin(x, d)
    
    TsIr(x, d): Rolling Information ratio over past d days
    
    TsVar(x, d): Rolling variance of x over the past d days
    
    TsSkew(x, d): Rolling skewness of x over the past d days
    
    TsKurt(x, d): Rolling kurtosis of x over the past d days
    
    TsMed(x, d): Rolling median of x over the past d days
    
    TsMad(x, d): Rolling median absolute deviation over the past d days
    
    TsRank(x, d): Time-series rank of x over the past d days
    
    TsDelta(x, d): Today's value of x minus the value of x d days ago
    
    TsRatio(x, d): Today's value of x divided by the value of x d days ago
    
    TsPctChange(x, d): Percentage change in x over the past d days
    
    TsWMA(x, d): Weighted moving average over the past d days with linearly decaying weights.
    
    TsEMA(x, d): Exponential moving average of x with span d
    
    TsCov(x, y, d): Time-series covariance of x and y for the past d days
    
    TsCorr(x, y, d): Time-series correlation of x and y for the past d days
    
(END OF FEATURES AND OPERATORS DEFINITIONS)\

Examples of valid alpha expressions:

Div(Sub(\$open, \$close), Add(Sub(\$high, \$low), 0.001))
}
\end{tcolorbox}

\begin{tcolorbox}[
  enhanced,                   % 开启增强功能
  breakable,                  % 支持跨页／跨栏
  colback=gray!5,             % 背景色
  colframe=blue!60!black,     % 边框色
  boxrule=0.8pt,              % 边框粗细
  arc=2mm,                    % 圆角半径
  left=4pt, right=4pt,        % 内边距
  top=4pt, bottom=4pt,
  width=\linewidth,           % 当前栏宽
  title={\bfseries User Prompts },
  fonttitle=\bfseries
]
% 切换到等宽小号字体并启用两端对齐
{\ttfamily\small\justifying
\textbf{User prompts for the strategy agent:}

Your task is to generate a new expression based on the given expressions,  the given topic {topic}, the explanation of this expression, and the given generation traces, such that:

1. The new expression is valid(syntactically correct), and dimensionless (i.e. Dim(expr) = 0).

2. You can only use the features, constants and operators given above in the (FEATURES AND OPERATORS DEFINITIONS). DO NOT MODIFY the name of any features,operators. All the operators should be used the same as their original definitions, ie, the number and type of arguments should be the same as defined.

3. As for constants, when you use it in rolling calculations, it should be an integer annoted by \%d ; when you use it in arithmetic calculations, it should be float numbers listed in 0.0001 0.01, 0.0, 1.0, 2.0. Do NOT use other constants. DO NOT use scientific notation.

4. Besides the Ref() operator, the constants used in rolling calculations should be the same, use "\%d" for all of them.

5. You should read the original expressions carefully, and try to understand its semantic meaning in quantitative finance. Then, you should try your best to think how to express the core of meaning in a different way, or generate new insights inspired by it. Then, you can generate a new expression based on your understanding.

6. If the trace is not empty, then it contains the generation optimization steps from the eariliest to the latest. You should learn how the expressions are optimzed before, and then generate a new expression based on the original expressions and the generation traces.

Given the original expresssion, you should ONLY and STRICTLY output a JSON object which contains the following contents:

\{{

  "strategies": ["the newly generated strategies for factor generation.The length of expressions should be \{{num}\}."],
  
\}}

Given expressions: \{expressions\}

Given expression explanations: \{explanations\}

}
\end{tcolorbox}

\begin{tcolorbox}[
  enhanced,                   % 开启增强功能
  breakable,                  % 支持跨页／跨栏
  colback=gray!5,             % 背景色
  colframe=blue!60!black,     % 边框色
  boxrule=0.8pt,              % 边框粗细
  arc=2mm,                    % 圆角半径
  left=4pt, right=4pt,        % 内边距
  top=4pt, bottom=4pt,
  width=\linewidth,           % 当前栏宽
  title={\bfseries User Prompts },
  fonttitle=\bfseries
]
% 切换到等宽小号字体并启用两端对齐
{\ttfamily\small\justifying
\textbf{User prompts for the execution agent:}

Your task is to generate a new expression based on the given expressions,  the given topic {topic}, the explanation of this expression, and the given generation traces, such that:

1. The new expression is valid(syntactically correct), and dimensionless (i.e. Dim(expr) = 0).

2. You can only use the features, constants and operators given above in the (FEATURES AND OPERATORS DEFINITIONS). DO NOT MODIFY the name of any features,operators. All the operators should be used the same as their original definitions, ie, the number and type of arguments should be the same as defined.

3. As for constants, when you use it in rolling calculations, it should be an integer annoted by \%d ; when you use it in arithmetic calculations, it should be float numbers listed in 0.0001 0.01, 0.0, 1.0, 2.0. Do NOT use other constants. DO NOT use scientific notation.

4. Besides the Ref() operator, the constants used in rolling calculations should be the same, use "\%d" for all of them.

5. You should read the original expressions carefully, and try to understand its semantic meaning in quantitative finance. Then, you should try your best to think how to express the core of meaning in a different way, or generate new insights inspired by it. Then, you can generate a new expression based on your understanding.

6. If the trace is not empty, then it contains the generation optimization steps from the eariliest to the latest. You should learn how the expressions are optimzed before, and then generate a new expression based on the original expressions and the generation traces.

7. The new expression should be different from all the given expressions, and should be novel and non-trivial, but it can share some common parts with them.

8. The new expression should be related to the given topic {topic}, and you should not generate expressions that are totally irrelevant to the topic.(EG: generate a expression about corr between price and volume when the topic is about volatility. Generate a expression containing rolling opertations when the topic is about interday prices, etc.)

9. Give \{num\} new expressions with different modification strateigies, with expressions soundness and explainable. The {num} expressions should be different, low correlated, looking semantically different, or using totally diiferent ways to specific semantics(For example, cross sectional(rank op) vs cross time-series(Ts op) or combine them; different statistical measures(Try to use new sound and interpretable operators that rarely or not exist in given expressions, traces and expressions generated before.); different but related semantic meanings, etc.). and each of them should be valid and dimensionless.

10. While maintaining relevance to the overarching topic\{topic\} and interpretability, you are encouraged to use as diverse a range of operators / features as possible across different expression outputs.

11. In each generation, there is no need to make the expression more complex, sometimes simplification is also a good way to generate novel expressions.

12. After you generate each expressions, check whether it is valid,some INVALID operations are:

   12.1 Modify the name of operators or features. Eg: using "*" instead of "Mul", using "closing\_price" instead of "\$close", using "TSMAD" instead of "TsMad", etc.
   
   12.2 The number of operands are not correct. Eg: Add(x), Div(x,y,z), Missing rolling window \%d in rolling operations(TsSkew, TsIr), etc.
   
   12.3 When using constant in algorithmic calculations, using integer instead of float, or vice versa.
   
   12.4 Use scientific notation in constants. You should transfer into float eg: 1e-4 -> 0.0001
   
   12.5 Using constants other than those mentioned in 3.
   
   12.6 Using the operators and features not mentioned in FEATURES AND OPERATORS  DEFINITIONS.
   
   All the mentioned above in 12 are INVALID operations. Try your best to AVOID / FIX Them.
   
Given the original expresssion, you should ONLY and STRICTLY output a JSON object which contains the following contents:

\{{

  "expressions": ["the newly generated expressions as a string list. Each elememnt should contain a valid and dimensionless expression.The length of expressions should be \{num\}."],
  "expressions\_fixed": ["for each generated expression, if there is any invalid operations mentioned above in 12, you should fix them and give the fixed expression here, otherwise just give the original generated expression here. The length of expressions\_fixed should be \{num\}. Do not  modify the semantics of the original generated expressions."],
  "explanations": ["brief explanations of  new expressions, should be the meaning of your given expression. The length of explanations should be {num}. And each explanation should correspond to the expression in the same index in the expressions list." ]
  
}\}

Given expressions: \{expressions\}

Given expression explanations: \{explanations\}

Generation traces: \{traces\}

Given Stratgies: \{straties\}

Do not output anything else other than the JSON object, including code block markers like ``` or ```json.

}
\end{tcolorbox}

\paragraph{Details about the validator agent}
A candidate factor $F_c$ is admitted to the pool only if it passes a check. The design of this check embodies a trade-off: we accept a factor if it either significantly improves upon existing ideas (exploitation) or introduces a valuable new dimension to our pool (exploration). Formally, admission requires:
\begin{equation*}
\begin{aligned}
    \Big( & \text{Qual}(F_c) > \tau_q \land\text{Gain}(F_c, F_p) > 0 \Big) 
      \lor \Big( \text{Qual}(F_c) > \tau_q \land \text{Corr}(F_c, \mathcal{F}) < \tau_d \Big)
\end{aligned}
\end{equation*}
where $\text{Qual}(\cdot)$ is a measure of predictive power implemented by the absolute value of ICIR, $\text{Gain}(\cdot, \cdot)$ measures the percentage improvement in $\text{Qual}(\cdot)$, and $\text{Corr}(\cdot, \cdot)$ measures the novelty of the candidate relative to the existing pool by calculating maximum abs value of  Pearson correlation coefficient between $F_c$ and each factor in $\mathcal{F}$. The $\tau_q$ is set to 0.10 and $\tau_d$ is set to 0.70. Upon admission, the DAG is updated with the new factor and a corresponding edge.

\subsection{Details about features and operators}
\label{appendix::detail:featop}
The table below summarizes all available features and operators we use in \MYMETHOD~ and all baselines. 
\begin{table*}[htbp]
\label{tab::3}
\centering
\caption{
    Raw features and operators. F: base market features; U/B: unary/binary operators; CS: cross-sectional operation (within-day across assets); TS: time-series rolling operation. The lookback length $d$ denotes the past $d$ trading days.
}
\begin{tabularx}{\textwidth}{@{} l l >{\raggedright\arraybackslash}X @{}}

\toprule
\textbf{Name} & \textbf{Type} & \textbf{Description} \\
\midrule
\multicolumn{3}{@{}l}{\textit{\textbf{Base Market Features}}} \\
Open    & F     & Opening price \\
Close   & F     & Closing price \\
High    & F     & Daily highest price \\
Low     & F     & Daily lowest price \\
Vwap    & F     & Daily average price, weighted by the volume of trades at each price \\
Volume  & F     & Trading volume (number of shares) \\
\midrule
\multicolumn{3}{@{}l}{\textit{\textbf{Unary and Cross-Sectional Operators}}} \\
Abs     & U     & Absolute value of the input \\
Slog1p  & U     & Signed log transform: $\text{sign}(\text{input}) \times \log(1 + |\text{input}|)$ \\
Inv     & U     & Reciprocal of the input; add $\epsilon$ to avoid division by zero \\
Sign    & U     & Sign of the input, returning -1, 0, or 1 \\
Log     & U     & Natural logarithm of the input; add $\epsilon$ for numerical stability \\
Rank    & U-CS  & Cross-sectional rank normalization within a day, mapped to the range [0, 1] \\
\midrule
\multicolumn{3}{@{}l}{\textit{\textbf{Binary Operators}}} \\
Add     & B     & Element-wise addition of two inputs \\
Sub     & B     & Element-wise subtraction: first minus second \\
Mul     & B     & Element-wise multiplication \\
Div     & B     & Element-wise division; add a small constant to the denominator for stability \\
Pow     & B     & Element-wise power: raise the first input to the power of the second \\
Greater & B     & Element-wise comparison: 1 if first input is greater than second, else 0 \\
Less    & B     & Element-wise comparison: 1 if first input is less than second, else 0 \\
GetGreater & B     & Element-wise comparison and retrieval: Retrieving the larger element \\
GetLess    & B     &  Element-wise comparison and retrieval: Retrieving the smaller element\\
\midrule
\multicolumn{3}{@{}l}{\textit{\textbf{Time-Series Operators}}} \\
Ref         & U-TS  & Lag operator: the value from $d$ days ago \\
TsMean      & U-TS  & Rolling mean over the past $d$ days \\
TsSum       & U-TS  & Rolling sum over the past $d$ days \\
TsStd       & U-TS  & Rolling standard deviation over the past $d$ days \\
TsIr        & U-TS  & Rolling information ratio over the past $d$ days \\
TsMinMaxDiff& U-TS  & Rolling range over the past $d$ days (rolling max minus rolling min) \\
TsMaxDiff   & U-TS  & Current value minus the rolling max over the past $d$ days \\
TsMinDiff   & U-TS  & Current value minus the rolling min over the past $d$ days \\
TsVar       & U-TS  & Rolling variance over the past $d$ days \\
TsSkew      & U-TS  & Rolling skewness over the past $d$ days \\
TsKurt      & U-TS  & Rolling kurtosis over the past $d$ days \\
TsMax       & U-TS  & Rolling maximum over the past $d$ days \\
TsMin       & U-TS  & Rolling minimum over the past $d$ days \\
TsMed       & U-TS  & Rolling median over the past $d$ days \\
TsMad       & U-TS  & Rolling median absolute deviation over the past $d$ days \\
TsRank      & U-TS  & Rolling rank of the current value within the past $d$ days, mapped to [0, 1] \\
TsDelta     & U-TS  & Change over $d$ days: current value minus the value $d$ days ago \\
TsDiv       & U-TS  & Ratio over $d$ days: current value divided by the value $d$ days ago \\
TsPctChange & U-TS  & Percentage change over the past $d$ days \\
TsWMA       & U-TS  & Linearly decaying weighted moving average over the past $d$ days \\
TsEMA       & U-TS  & Exponential moving average with a decay over the past $d$ days \\
TsCov       & B-TS  & Rolling covariance between two inputs over the past $d$ days \\
TsCorr      & B-TS  & Rolling Pearson correlation between two inputs over the past $d$ days \\
\bottomrule
\end{tabularx}
\end{table*}

\begin{figure}[htbp]
    \centering
    \includegraphics[width=0.7\linewidth]{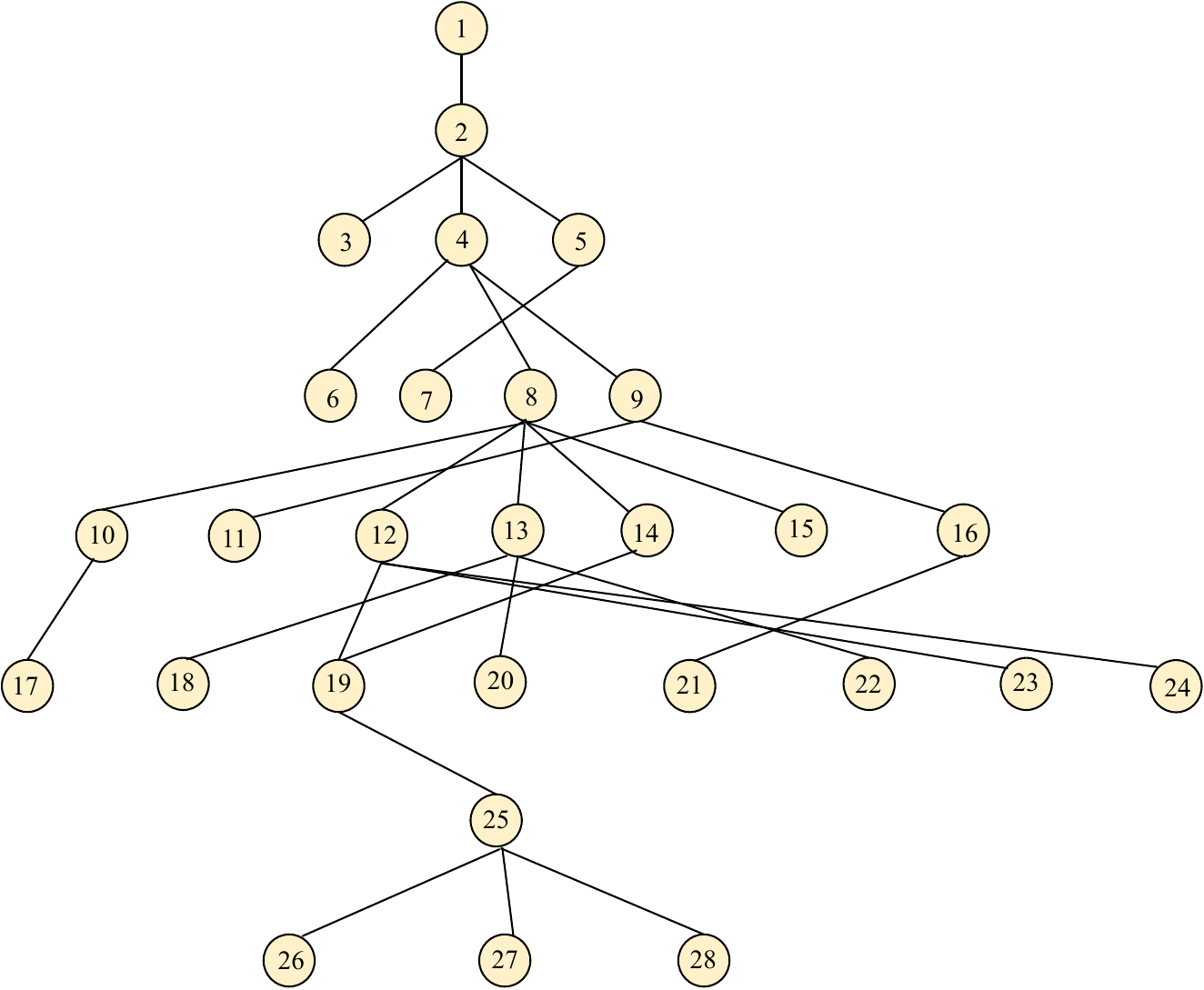}
    \caption{The topological structure of mining factors on the CSI 300 which derives from interday price movements. For more details, please refer to Table~\ref{table:factors}. The node indices in the figure correspond to the Factor ID in the table.
}
    \label{fig:case_study}
\end{figure}

% 请确保在您的文档序言部分包含了以下宏包
% \usepackage{tabularx}
% \usepackage{booktabs}
% \usepackage{amsmath} % 推荐使用，用于更复杂的数学表达式

% 请在您的文档导言区确保已经包含了以下宏包:
% \usepackage{booktabs}   % 用于 \toprule, \midrule, \bottomrule
% \usepackage{xltabular}  % 用于跨页表格，并支持 X 列类型

% \newcolumntype{L}{>{\raggedright\arraybackslash}X} % 这句可以保留在导言区或表格前

\newcolumntype{L}{>{\raggedright\arraybackslash}X}
\begin{xltabular}{\textwidth}{@{} l L L l @{}} % <-- 1. 环境从 tabularx 改为 xltabular, 并移除了 table* 环境

% --- 表格标题和标签 ---
\caption{Details about all factors in Figure~\ref{fig:case_study}. We report the factor ID, factor expressions and descriptions and their ICIR on the test period.}
\label{table:factors} \\ % <-- 2. 标题和标签移到环境内部

% --- 定义每一页的表头 ---
% 2.1. 第一页的表头
\toprule
\textbf{Factor ID} & \textbf{Expression} & \textbf{Description} & \textbf{ICIR} \\
\midrule
\endfirsthead

% 2.2. 后续页面的表头 (和第一页一样，但可以加上 "continued" 提示)
\caption[]{(continued)} \\ % "[]" 使得这个续表标题不会出现在目录的 "List of Tables" 中
\toprule
\textbf{Factor ID} & \textbf{Expression} & \textbf{Description} & \textbf{ICIR} \\
\midrule
\endhead

% 2.3. 表格最后一页的页脚 (在这里就是表格的结束线)
\bottomrule
\endlastfoot

% --- 表格主体内容 (和原来完全一样) ---
1 & Div(Sub(Less(\$open, \$close), \$low), \$open)
  & This expression measures the normalized daily price reversal signal by calculating the difference between a binary indicator of positive close-open return (1 if close $>$ open, else 0) and the low price, then scaling it by the opening price, which is useful for identifying potential mean-reversion opportunities.
  & 0.2391\\

2 & TsCorr(Sub(\$close, \$open), Sub(\$low, TsMin(\$low, 5)), 5)
  & Correlates the daily return with the deviation of low price from its 5-day minimum, assessing co-movement in interday price behavior & 0.2652 \\

3 & TsCorr(Sub(\$close, \$open), TsRank(Sub(\$low, TsMin(\$low, 5)), 5), 5)
  & Correlates the daily return with the time-series rank of the low price deviation from its 5-day minimum, evaluating if returns associate with the relative extremity of low prices. & 0.0947 \\

4 & TsCorr(TsDelta(\$close, 1), TsMinDiff(\$low, 5), 5)
  & Correlates the one-day price change (close delta) with the difference between current low and its 5-day minimum, evaluating momentum and support level interactions. & 0.0611 \\

5 & TsCorr(Sub(\$close, \$open), Div(Sub(\$low, TsMin(\$low, 5)), TsStd(\$low, 5)), 5)
  & Correlates the daily return with the normalized deviation of low price from its 5-day minimum (scaled by volatility), assessing if returns relate to standardized low price extremes. & 0.1179 \\

6 & TsCorr(TsPctChange(\$close, 5), TsMinDiff(\$low, 5), 5)
  & Correlates the one-day percentage change in closing price with the deviation of low from its 5-day minimum, linking momentum to support breakout signals. & 0.1319 \\

7 & TsCorr(Sub(\$close, \$open), TsRank(Div(Sub(\$low, TsMin(\$low, 5)), TsStd(\$low, 5)), 5), 5)
  & Correlates the daily return with the time-series rank of the normalized deviation of low price from its 5-day minimum, assessing if returns relate to the relative standing of standardized low price extremes over the past 5 days. & 0.1614 \\

8 & TsCorr(TsDelta(\$close, 5), TsMaxDiff(\$low, 5), 5)
  & Correlates the one-day change in closing price with the deviation of low from its 5-day maximum, evaluating momentum against resistance rather than support levels. & 0.1507 \\

9 & TsCorr(TsDelta(\$close, 5), TsDelta(\$low, 5), 5)
  & Correlates the one-day change in closing price with the one-day change in low price, capturing co-movement in interday price dynamics. & 0.1998 \\

10 & TsCorr(TsPctChange(\$close, 5), TsPctChange(\$low, 5), 5)
  & Correlates the percentage change in closing price with the percentage change in low price, assessing relative sensitivity in interday price movements. & 0.1632 \\

11 & TsCorr(TsDelta(\$close, 5), Sub(TsMax(\$high, 5), \$close), 5)
  & Correlates the one-day change in closing price with the gap between the current close and the maximum high over a period, focusing on breakout potential from resistance. & 0.2639 \\

12 & TsCorr(TsDelta(\$close, 5), TsDelta(\$vwap, 5), 5)
  & Correlates the one-day change in closing price with the one-day change in volume-weighted average price, assessing alignment between closing trends and average traded price movements. & 0.1210 \\

13 &TsCorr(TsPctChange(\$close, 5), TsPctChange(\$vwap, 5), 5)
  & Correlates the percentage change in closing price with the percentage change in volume-weighted average price, evaluating price efficiency and market impact. & 0.0719 \\

14 &TsCorr(TsDelta(\$close, 5), TsDelta(Sub(\$high, \$low), 5), 5)
  & Correlates the one-day change in closing price with the one-day change in the daily range (high minus low), capturing how closing momentum relates to daily volatility changes. & 0.2813 \\

15 &TsCorr(TsDelta(\$close, 5), TsDelta(\$volume, 5), 5)
  & Correlates the one-day change in closing price with the one-day change in trading volume, capturing price-volume dynamics in interday movements. & 0.2639 \\

16 &TsCorr(TsDelta(\$close, 5), TsPctChange(\$low, 5), 5)
  & Correlates the one-day change in closing price with the percentage change in low price over d days, measuring momentum against relative low price shifts. & 0.0935 \\

17 &TsCorr(TsPctChange(\$close, 5), Sub(\$high, \$low), 5)
  & Correlates the percentage change in closing price with the daily range (high minus low), evaluating how closing trends relate to intraday volatility. & 0.2746 \\
  \bottomrule
  
18 &TsCorr(Sub(\$close, Ref(\$close, 1)), Sub(\$vwap, Ref(\$vwap, 1)),5)
  & Correlates the absolute daily change in closing price with the absolute daily change in VWAP, capturing co-movement in raw price adjustments rather than percentage changes. & 0.1252 \\
19 &Rank(TsCorr(TsDelta(\$close, 5), TsDelta(\$vwap, 5), 5))
  & Ranks the correlation between daily changes in close and VWAP cross-sectionally, identifying stocks with strongest alignment between closing trends and average traded prices. & 0.1169 \\
20 &TsCorr(TsSkew(\$close, 5), TsKurt(\$vwap, 5), 5)
  &Correlates the rolling skewness of closing prices with the rolling kurtosis of VWAP, measuring the association between price distribution asymmetry and tail risk in market pricing. & 0.1218 \\
21 &TsCorr(TsPctChange(\$close, 5), TsDelta(\$low, 5), 5)
  &Correlates the percentage change in closing price over d days with the one-day change in low price, evaluating interday momentum and low price dynamics. & 0.1452 \\
22 &TsCorr(TsDelta(\$close, 5), TsPctChange(\$volume, 5), 5)
  &Correlates one-day change in closing price with percentage change in volume, assessing relationship between price movements and trading activity shifts over interday periods. & 0.1760 \\
23 &TsCorr(TsPctChange(\$close, 5), TsPctChange(\$volume, 5), 5)
  &Correlates the percentage change in closing price with the percentage change in volume, evaluating the relationship between interday price momentum and trading activity intensity. & 0.2681 \\
24 &Mul(TsCorr(TsDelta(\$close, 5), TsDelta(\$vwap, 5), 5), TsSkew(TsDelta(\$close, 5), 5))
  &Multiplies the correlation between daily changes in close and VWAP with the skewness of close price changes, capturing both alignment and asymmetry in interday price movements. & 0.1015 \\
25 &Rank(TsCorr(TsPctChange(\$close, 5), TsPctChange(\$vwap, 5), 5))
  &Ranks the correlation between percentage changes in close and VWAP cross-sectionally, assessing alignment in relative price movements rather than absolute changes.& 0.0974 \\
26 &Rank(TsCorr(TsPctChange(\$close, 5), TsPctChange(\$low, 5), 5))
  &Ranks the correlation between percentage changes in close and low prices cross-sectionally, evaluating alignment in downward price momentum movements.& 0.1731 \\
27 &Rank(TsCorr(TsPctChange(\$close, 5), TsDelta(\$volume, 5), 5))
  &Ranks the correlation between percentage changes in close price and absolute changes in volume, assessing price-volume relationship in interday dynamics.& 0.2765 \\
28 &Sub(Rank(TsCorr(TsPctChange(\$close, 5), TsPctChange(\$vwap, 5), 5)), Rank(TsCorr(TsPctChange(\$close, 5), TsDelta(\$volume, 5), 5)))
  &Subtracts the ranked correlation of price change with volume change from that with VWAP change, isolating VWAP-specific interday price dynamics.& 0.2281 \\
\end{xltabular}